\documentclass[11pt]{article}

\usepackage[final]{acl}

\usepackage{latexsym}
\usepackage{xcolor}
\usepackage{mdframed}
\usepackage{float}
\usepackage{fontspec}
\usepackage{xeCJK}
\setCJKsansfont{FandolHei-Regular.otf}[
  BoldFont = FandolHei-Bold.otf
]

\newfontfamily\cyrfont{FreeSerif.otf}
\newfontfamily\arabfont{FreeSerif.otf}
\newfontfamily\thaifont{FreeSerif.otf}
\newfontfamily\greekfont{FreeSerif.otf}

\usepackage[utf8]{inputenc}

\usepackage{microtype}

\usepackage{inconsolata}

\usepackage{graphicx}
\usepackage{booktabs}
\usepackage{multirow}
\usepackage{dblfloatfix}
\usepackage{enumitem}
\usepackage{amssymb} 
\usepackage{amsmath} 

\title{Tracing the Latent Threads: A Mechanistic Study of How LLMs Represent and Operationalize Race and Ethnicity Cues}

\author{
  Shiyue Hu\textsuperscript{1,2} \quad
  Ruizhe Li\textsuperscript{3,*} \quad
  Yanjun Gao\textsuperscript{1,*} \\
  \textsuperscript{1}University of Colorado Anschutz \\
  \textsuperscript{2}University of Colorado Boulder \\
  \textsuperscript{3}University of Birmingham \\
  \texttt{shiyue.hu@colorado.edu, r.li.7@bham.ac.uk, yanjun.gao@cuanschutz.edu} \\
  \textsuperscript{*}Co-senior authors
}

\begin{document}
\maketitle
\begin{abstract}
Large language models (LLMs) increasingly operate in high-stakes settings where demographic attributes such as race and ethnicity may be explicitly stated or implicitly suggested through textual cues. However, existing studies primarily document outcome-level disparities, offering limited insight into internal mechanisms underlying these effects. We present a mechanistic study of how race and ethnicity cues are represented and operationalized within LLMs. Using two publicly available datasets spanning toxicity-related generation and clinical narrative understanding tasks, we analyze three open-source models with a reproducible interpretability pipeline combining probing, neuron-level attribution, and targeted intervention. We find that sensitivity to demographic cues is distributed across internal units and varies substantially across models. These units often align with entangled semantic facets, including explicit group labels, geography, language, culture, and associations related to stereotypes. Interventions on selected units can change some biased prediction patterns, but substantial residual effects remain, suggesting that effective mitigation requires understanding distributed, task-specific mechanisms rather than manipulating a small set of identified neurons alone.

\end{abstract}

\section{Introduction}\footnote{We utilized ChatGPT to identify and translate non-English tokens presented in this paper.}
Large language models (LLMs) are increasingly used in high-stakes domains such as healthcare, where demographic attributes (e.g., race, ethnicity, gender) may be explicitly stated or implicitly inferred from text. Prior work shows that LLMs can condition their outputs on demographic information even when it is not task-relevant~\cite{zack2024assessing,kim-etal-2023-race,fraser-kiritchenko-2024-examining,zhao-etal-2025-explicit}, and can therefore induce misattribution in model output with undesirable or biased behavior~\cite{demchak2024assessing,yan2026spurious,levartovsky2025sociodemographic,li2026attributing,zack2024assessing}. 

Most prior studies on demographic bias focus on outcome-level effects, evaluating disparities in generated responses, accuracy, calibration, or toxicity scores across demographic groups~\cite{tan-lee-2025-unmasking,hartvigsen-etal-2022-toxigen,guan-etal-2025-saged,wang-etal-2025-fairness}. While these analyses are essential for documenting harm, they treat LLMs as black boxes, offering limited insight into whether demographic attributes are reflected in high-level semantic features, task-relevant representations, or spurious shortcuts during prediction. In parallel, recent works in mechanistic interpretability demonstrated how LLMs represent demographic information and manipulated LLM internal states to ensure fairness ~\cite{yu2025understanding,ahsan2025can,karvonen2025robustly}, yet these tools have rarely been applied to demographic bias in a systematic and task-diverse manner.  

A central challenge is that demographic attributes interact with language in complex ways. In many real-world settings, demographic information may be explicitly stated (e.g., ``a Black patient'', ``a Hispanic speaker'') or implicitly conveyed through linguistic, cultural, or geographical cues, i.e., ``proxy'' cues. Moreover, the same demographic signal can have qualitatively different effects depending on task: it may alter predicted medical risk in a clinical scenario, while simultaneously modulating perceived toxicity, credibility, or intent in open-ended generation. Existing evaluations typically isolate a single task or domain~\cite{zack2024assessing,levartovsky2025sociodemographic}, making it difficult to assess whether demographic sensitivity reflects general representational mechanisms or task-specific heuristics.

In this work, we investigate how demographic attributes influence LLM behavior, with a focus on mechanistic explanations rather than surface-level disparities. We examine \textit{race} and \textit{ethnicity} as commonly occurring coarse-grained categories (e.g. White, Black, Asian, Hispanic and Latino) as they appear in the studied datasets, rather than attempting to model the full sociological complexity of these constructs. Using two publicly available datasets, we study: 1) toxicity-related generation tasks~\cite{hartvigsen-etal-2022-toxigen}, where the same attributes may alter the likelihood, tone, or framing of model outputs, and 2) clinical narrative tasks~\cite{beardontwalk2024creact}, where the same attributes appear through explicit or indirect cues in medical text and modulate model behavior despite identical clinical evidence.  

We adopt a mechanistic interpretability (MI) framework to study how lexical cues associated with race and ethnicity are represented and propagated within three open-source LLMs that are widely used: Qwen2.5-7B~\cite{qwen2.5}, Mistral-7B~\cite{jiang2023mistral}, and Llama-3.1-8B~\cite{grattafiori2024llama3}. Our contributions are threefold: 
\noindent
\begin{itemize}[leftmargin=*, nosep]
    \item a reproducible MI \textbf{pipeline} that combines multi-class probing, neuron-level attribution, and targeted intervention to identify internal units associated with demographic attributes and to examine their functional relevance across tasks. The proposed framework is applicable to other social variables beyond race and ethnicity. 
    \item a fine-grained \textbf{characterization}  of internal features associated with race and ethnicity cues across LLMs, revealing the distributed nature of demographic information and model-specific emphasis on semantic facets such as geography, language, culture, or historical context. 
    \item a mechanistic \textbf{analysis} of how demographic representation influences model behaviors. Internal features aligned with sensitive or harmful stereotype-related concepts in the pretraining data are unevenly activated by direct and indirect demographic cues.  
\end{itemize}

Our findings show that while features associated with race and ethnicity cues can be identified and analyzed at the neuron level, their associations with biased model behavior persist even when highly active neurons are sign-flipped. This indicates that biased behavior in LLMs cannot be fully explained or controlled by manipulating a small set of identifiable neurons alone.


\section{Related Work}
\begin{figure}[tb]
    \centering
    \includegraphics[width=\columnwidth]{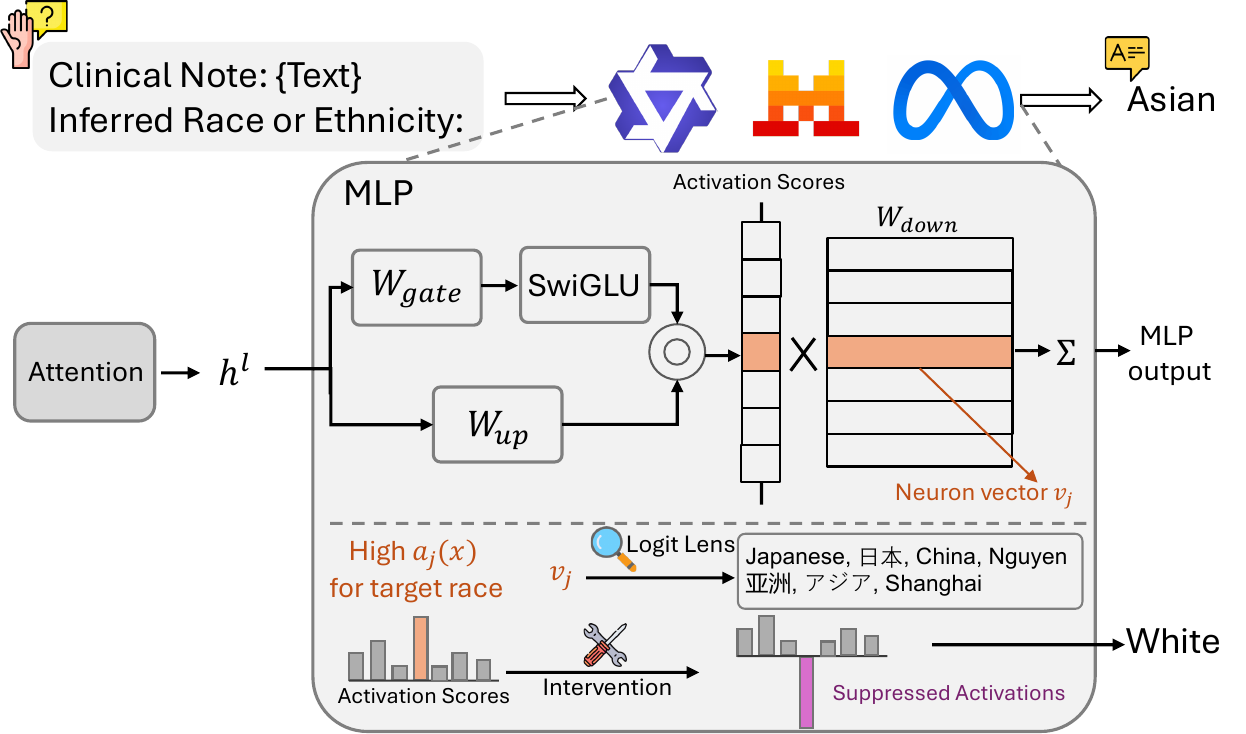}
    \vspace{-.16in}
    \caption{\small With MLP, we locate neurons associated with race and ethnicity cues and inspect them via Logit Lens. For neurons with high activation on target race and ethnicity cues, we intervene on their activation values to test their causal influence on model behavior.}
    \vspace{-1.5em}
    \label{fig:mlp_framework}
\end{figure}
\noindent \textbf{Mechanistic Interpretability of Bias in LLMs.}
Recent works in mechanistic interpretability have begun to locate where demographic information is represented in LLMs. Our approach of using probe-based neuron localization aligns with emerging research in this space.~\citet{yu2025understanding} utilized neuron editing to understand and mitigate gender bias, identifying specific ``gender neurons'' within the MLP layers. In the medical domain,~\citet{ahsan2025elucidating} investigated the mechanisms of demographic bias specifically for healthcare tasks, suggesting that certain internal representations are disproportionately sensitive to racial identifiers. More recently,~\citet{ahsan2025can} explored the use of Sparse Autoencoders (SAEs) to reveal clinical racial biases. Our work extends these findings by demonstrating that neurons associated with race and ethnicity cues are not only present in general datasets but are consistently activated and influential on domain-specific text. 

\noindent \textbf{Internal Bias Mitigation and Mechanistic Steering within LLMs.}
Beyond identification, recent work focuses on manipulating internal model states to ensure fairness.~\citet{zhou2024unibias} proposed the UniBias framework, which mitigates bias by manipulating attention heads and MLP components. For real-time applications,~\citet{li2025fairsteer} introduced \textit{FairSteer}, a dynamic activation steering method that adjusts model behavior during inference.~\citet{karvonen2025robustly} further demonstrated that interpretability-based interventions can improve fairness more robustly than traditional fine-tuning in realistic settings. Our methodology contributes to this line of work by using targeted intervention, specifically sign-flipping and amplification, to test whether identified race-associated neurons causally influence biased prediction patterns. This builds upon the ``context-aware'' fairness frameworks suggested by~\citet{nadeem2025context}, ensuring that mitigation is grounded in the semantic understanding of the racial directions we extract.

\noindent \textbf{Racial Bias in Clinical LLMs.}
Extensive research found that LLMs inherit and propagate racial biases when applied to clinical decision support.~\citet{zhang2023chatgpt} demonstrated that ChatGPT exhibits disparate treatment recommendations for acute coronary syndrome based on racial and gender cues. Similarly,~\citet{zack2024assessing} evaluated GPT-4, finding that model frequently perpetuates harmful stereotypes that could lead to inequitable health outcomes.~\citet{poulain2024bias} further expanded this analysis across various clinical decision-support tasks, highlighting that bias patterns are not idiosyncratic but systematic across model families. While these studies establish the existence of bias, they largely treat model as a black box. Our work seeks to uncover internal mechanisms driving these disparate outputs.

\section{Background}

\noindent \textbf{MLP Layers and Neuron Activation.}
Modern Transformer-based LLMs process information through a \textit{residual stream}. 
In this framework, the residual stream acts as a communication channel, while MLP layers function as key-value memories that store and inject factual associations into the stream~\citep{geva-etal-2021-transformer}.
Contemporary models like Llama 3.1, Mistral, and Qwen 2.5 utilize the \textit{SwiGLU} gated architecture~\citep{shazeer2020glu}. The output of an MLP block with input $x$ is defined as:
\begin{equation}\small
\text{MLP}(x) = \left( \text{SwiGLU}(x W_{\text{gate}}) \odot (x W_{\text{up}}) \right) W_{\text{down}}
\vspace{-.5em}
\end{equation}
where $\odot$ is the element-wise product. We define an individual \textit{neuron} $j$ as the $j$-th element of the intermediate gated state. The total MLP output is the sum of these neurons' contributions:
\begin{equation}\small
\text{MLP}(x) = \sum_{j=1}^{d_{mlp}} a_{j}(x) \cdot v_{j}
\vspace{-.5em}
\end{equation}
where $d_{\mathrm{mlp}}$ denotes the intermediate MLP width, $a_j(x)$ is the activation score of neuron $j$ after the gated projection, and $v_j$ is the $j$-th output vector from $W_{\mathrm{down}}$. Our method specifically probes these \textit{output vectors} $v_j$ to locate racial information.

\noindent \textbf{Logit Lens.}
To interpret high-dimensional vectors in the residual stream or neuron output vectors $v_j$, we use \textit{Logit Lens}~\citep{nostalgebraist2020logitlens}. This technique projects a vector $h$ directly into vocabulary space using model's unembedding matrix $W_U$: $\text{logits} = h W_U$. By inspecting top-ranked tokens in the resulting distribution, we can decode the semantic concepts associated within specific neurons.

\section{Methodology}
\label{sec:methodology}

We propose a mechanistic interpretability framework to determine \textit{where} and \textit{how} race-associated information is represented within LLMs. Our approach progresses from identifying global race directions via multi-class probing to identifying the specific neurons responsible for these representations.

\subsection{Locating Race Directions via Multi-Class Probing}

To extract race/ethnicity representations, we train linear probes $W_{\text{Race}}$ to classify race/ethnicity category membership for each model. The probe is trained on the final-layer residual stream $\bar{h}^{L-1}$, averaged across all token positions:
\begin{equation}
\small
P(\text{race} = c \mid \bar{h}^{L-1}) = \text{softmax}(W_{\text{Race}}^\top \bar{h}^{L-1} + b)_c
\end{equation}
where $W_{\text{Race}} \in \mathbb{R}^{d \times |\mathcal{C}|}$ is the learned probe matrix, $b$ is bias vector, and $\mathcal{C}$ denotes the set of race/ethnicity categories. Each column $w_c$ of $W_{\text{Race}}$ represents \textit{race direction} for group $c$ in model's representation space.

\subsection{From Race Directions to Neurons}

Having identified the global race directions $w_c$, we locate the MLP neurons that write to these directions, motivated by prior work showing MLPs act as key–value memories \citep{geva-etal-2021-transformer}.

\noindent \textbf{Interpreting the Probe Direction.}
We first verify that our learned directions $w_c$ capture meaningful racial semantics. Using Logit Lens, we project each direction into vocabulary space via the model's unembedding matrix $W_U$:
\begin{equation}\small
z_{\text{probe}} = W_U w_c
\vspace{-.5em}
\end{equation}
Top-$k$ tokens ($k=20$) with the highest values in $z_{\text{probe}}$ serve as a semantic fingerprint for each racial group.

\noindent \textbf{Identifying Candidate Neurons.}
To locate neurons that write to race direction, we compute the cosine similarity between each MLP neuron's output vector $v_{j}^l$ at layer $l$ and the probe direction $w_c$:
\begin{equation}\small
\text{Score}(l, j) = \frac{w_c \cdot v_{j}^l}{\|w_c\| \|v_{j}^l\|}
\vspace{-.5em}
\end{equation}
All neurons in the final four MLP layers are ranked by this score, and the top 20 candidates are selected. Each candidate is selected by projecting its output vector into vocabulary space and inspecting top-20 tokens. Neurons are retained only if their tokens show clear alignment with the target racial group. 

\subsection{Validating Candidate Neurons via Activation Analysis and Intervention}

To evaluate whether identified neurons reflect meaningful racial information and influence model behavior, we design a two-stage validation pipeline.

\noindent \textbf{Activation Analysis.}
We measure how strongly each neuron group activates across different inputs. For a given input text, we extract the activation score of each candidate neuron during the forward pass. For ToxiGen, we average activations across all token positions; for C-REACT, where the model must produce a classification, we extract activations at the last input token before generation. We then compare activation patterns across racial groups: if neurons identified for a particular race show higher activation scores when processing text about that group compared to other groups, this provides evidence that these neurons selectively respond to race-related content.

\noindent \textbf{Targeted Intervention.}
To test whether identified neurons causally influence model outputs, we manipulate their activations during inference. We implement this using PyTorch forward hooks on the MLP $W_{\text{down}}$. During the forward pass, we intercept the activation score of each targeted neuron and force it into a strongly negative range. Specifically, we multiply positive activations by a negative factor (e.g., -5) and amplify negative activations by a positive factor (e.g., 5). This makes the neuron’s contribution subtractive rather than additive. This modified signal propagates through the remaining layers, allowing us to measure whether steering specific neurons changes model behavior and final outputs.

\section{Data and Models}
\label{sec:data}
\textbf{Datasets.}
We evaluate our framework on two domains to test the generalizability of race-associated mechanisms across different contexts.
\textbf{ToxiGen} \citep{hartvigsen-etal-2022-toxigen} is a large-scale machine-generated dataset of toxic statements about minority groups. We use the annotated subset containing 9.9k samples across 13 target groups. Each sample has a statement for a specific group (e.g., \textit{``asians are trying to take over the world with their karate skills''}).
From all groups, we consolidate related categories into four racial categories: Asian (asian, chinese), Black (black), Latino (latino, mexican), and Native American.
\textbf{C-REACT} (Contextualized Race and Ethnicity Annotations for Clinical Text \citep{beardontwalk2024creact}) provides race and ethnicity annotations for 17,281 sentences drawn from clinical notes in the MIMIC-III database. C-REACT contains real clinical text where race information appears in two forms: direct mentions that explicitly state race (e.g., \textit{``Pt is 42 yo AA female''}) and indirect mentions that imply race through associated attributes such as spoken language or country of origin (e.g., \textit{``Pt required a Spanish interpreter''}, \textit{``Pt is recently immigrated from France''}). 
C-REACT provides five racial categories: White, Black/African American (Black/AA), Asian, Native American or Alaska Native, and Native Hawaiian or Other Pacific Islander. However, the dataset is highly imbalanced: zero patients labeled as Native Hawaiian or Other Pacific Islander were found, and only three patients labeled as Native American or Alaska Native. We therefore use three racial categories with sufficient representation: White, Black/AA, and Asian. 

\noindent \textbf{Models.}
We study three instruction-tuned LLMs of comparable scale from different geographic and cultural training contexts: Llama-3.1-8B-IT \citep{grattafiori2024llama3} (US), Mistral-7B-IT-v0.3 \citep{jiang2023mistral} (France), and Qwen2.5-7B-IT \citep{qwen2.5} (China). This selection allows us to investigate whether models trained on data from different linguistic and cultural contexts develop different internal responses to race and ethnicity cues, given that conceptions of race and ethnicity vary across societies. 

\section{Experiments and Results}

\subsection{ToxiGen}
We first examine whether the learned race and ethnicity directions recover interpretable semantic facets of demographic cues in toxicity-related text.
Table~\ref{tab:probe_tokens} lists top tokens projected by each race direction. Across models, probes reach similar performance on ToxiGen (around 75\% accuracy/macro-F1; Appendix~\ref{app:probe_metrics}). These tokens capture various facets of racial representations, including geography, religion, demographic labels, and cultural terms. Across all three models, the learned directions identify tokens that align closely with the target race/ethnicity categories. This suggests that the learned probe directions capture demographic cue information in the residual stream.
\begin{table}[tb]
\centering
\small
\resizebox{\columnwidth}{!}{%
\begin{tabular}{llll}
\toprule
\textbf{Model} & \textbf{Group} & \textbf{Top tokens projected by probe} \\
\midrule
\multirow{3}{*}{Qwen2.5-7B} 
    & Asian & Asian, 亚洲, Chinese, CJK, 东亚 \\
    \cmidrule(lr){2-3}
    & Latino & Mex, Mexico \\
    \cmidrule(lr){2-3}
    & Native American & natives, native, Native, indigenous \\
\midrule
\multirow{5}{*}{Mistral-7B} 
    & Asian & Chinese, Asian, China, Korean, Taiwan \\
    \cmidrule(lr){2-3}
    & Black & black, African, Black \\
    \cmidrule(lr){2-3}
    & Latino & Mexico, Salvador, Colombia, Chile, Mexican \\
    \cmidrule(lr){2-3}
    & Native American & Indians, trib, Native, tribes, Indian \\
\midrule
\multirow{5}{*}{Llama-3.1-8B} 
    & Asian & Asian, Mandarin, CJK, asian, china \\
    \cmidrule(lr){2-3}
    & Black & Black, \_black, -black, .black, 黑 \\
    \cmidrule(lr){2-3}
    & Latino & Mundo, \_BORDER \\
    \cmidrule(lr){2-3}
    & Native American & Native, natives, Indians, indigenous, tribes \\
\bottomrule
\end{tabular}%
}
\vspace{-.05in}
\caption{\small Top tokens by race group across models (ToxiGen). \textbf{WARNING: Some tokens reflect harmful stereotypes.} Full results in Appendix~\ref{app:ToxiGen_tokens}.
{\scriptsize Translations: 亚洲 (Asia), 东亚 (East Asia), 黑 (Black).}}
\label{tab:probe_tokens}
\vspace{-1.1em}
\end{table}

\begin{table}[tb]
\centering
\resizebox{\columnwidth}{!}{%
\begin{tabular}{llll}
\toprule
\textbf{Model} & \textbf{Group} & \textbf{Neuron} & \textbf{Top tokens} \\
\midrule
\multirow{9}{*}{Qwen2.5-7B}
    & \multirow{2}{*}{Asian} 
        & MLP.v$^{28}_{13406}$ & Japanese, 日本, Japan, Tokyo \\
    &   & MLP.v$^{25}_{15029}$ & Chinese, China, 中国, Asian \\
    \cmidrule(lr){2-4}
    & \multirow{1}{*}{Black} 
        & MLP.v$^{27}_{2240}$ & black, 黑, Black, 黑色 \\
    \cmidrule(lr){2-4}
    & \multirow{2}{*}{Latino} 
        & MLP.v$^{28}_{4781}$ & Latin, 拉丁, latino, Latina \\
    &   & MLP.v$^{27}_{18125}$ & Spanish, Hispanic, Chile, Mexican \\
    \cmidrule(lr){2-4}
    & \multirow{2}{*}{Native Am.} 
        & MLP.v$^{25}_{3458}$ & native, Native, indigenous \\
    &   & MLP.v$^{25}_{11197}$ & colonial, colon, colony, imperial \\
\midrule
\multirow{9}{*}{Mistral-7B}
    & \multirow{1}{*}{Asian} 
        & MLP.v$^{32}_{4453}$ & Japanese, Korean, Taiwan, Asian \\
    \cmidrule(lr){2-4}
    & \multirow{2}{*}{Black} 
        & MLP.v$^{32}_{5923}$ & Black, black, Negro, African \\
    &   & MLP.v$^{30}_{12572}$ & Black, blacks, 黑, dark \\
    \cmidrule(lr){2-4}
    & \multirow{2}{*}{Native Am.} 
        & MLP.v$^{31}_{3440}$ & colonial, colon \\
    &   & MLP.v$^{29}_{12205}$ & native, ind, igenous \\
\midrule
\multirow{9}{*}{Llama-3.1-8B}
    & \multirow{2}{*}{Asian} 
        & MLP.v$^{32}_{5691}$ & Chinese, China, Beijing, 中国 \\
    &   & MLP.v$^{31}_{14299}$ & Li, yuan, Dong, Huang, Wang \\
    \cmidrule(lr){2-4}
    & \multirow{2}{*}{Black} 
        & MLP.v$^{30}_{7195}$ & Jamaica, Caribbean, Trinidad, Jazz \\
    &   & MLP.v$^{29}_{13826}$ & African, Afro, negro, blacks \\
    \cmidrule(lr){2-4}
    & \multirow{1}{*}{Latino} 
        & MLP.v$^{32}_{9242}$ & Spanish, Hispanic, Mexican, Argentine \\
    \cmidrule(lr){2-4}
    & \multirow{2}{*}{Native Am.} 
        & MLP.v$^{32}_{6893}$ & colon, colonial, colonization, colonies \\
    &   & MLP.v$^{31}_{1186}$ & native, Native, -native, natives \\
\bottomrule
\end{tabular}
}
\vspace{-.1in}
\caption{\small Top race-associated neurons identified via cosine similarity with probe directions. \textbf{WARNING: Some tokens reflect harmful stereotypes.} Full results in Appendix~\ref{app:ToxiGen_neurons}.
{\scriptsize Translations: 日本 (Japan), 中国 (China), 黑 (Black), 黑色 (Black color), 拉丁 (Latin).}}
\vspace{-2em}
\label{tab:neurons}
\end{table}

Table~\ref{tab:neurons} presents race-associated neurons identified within the final four MLP layers. These neurons reveal that race and ethnicity cues are associated with multiple semantic facets. Some neurons align with broad demographic terminology that directly names groups, such as Mistral-7B's $\text{MLP.v}^{32}_{5923}$ (\textit{Black, black, African}) and Asian neurons across all models (\textit{Asian, Chinese, Japanese}). Other neurons align with attributes that frequently co-occur with demographic labels in text. For example, Llama-3.1-8B's $\text{MLP.v}^{32}_{5691}$ links Asian identity to geographic terms (\textit{Chinese, China, Beijing}), while $\text{MLP.v}^{31}_{14299}$ captures Chinese last names (\textit{Li, yuan, Dong, Huang, Wang}). We also observe neurons whose top projected tokens reflect historically harmful associations. Native American neurons across all three models project to colonial terminology (\textit{colonial, colony, colonization}), and neurons associated with Black identity recover offensive racial terms that persist across models despite different training corpora.

\begin{figure*}[ht]
\centering
\includegraphics[width=\textwidth]{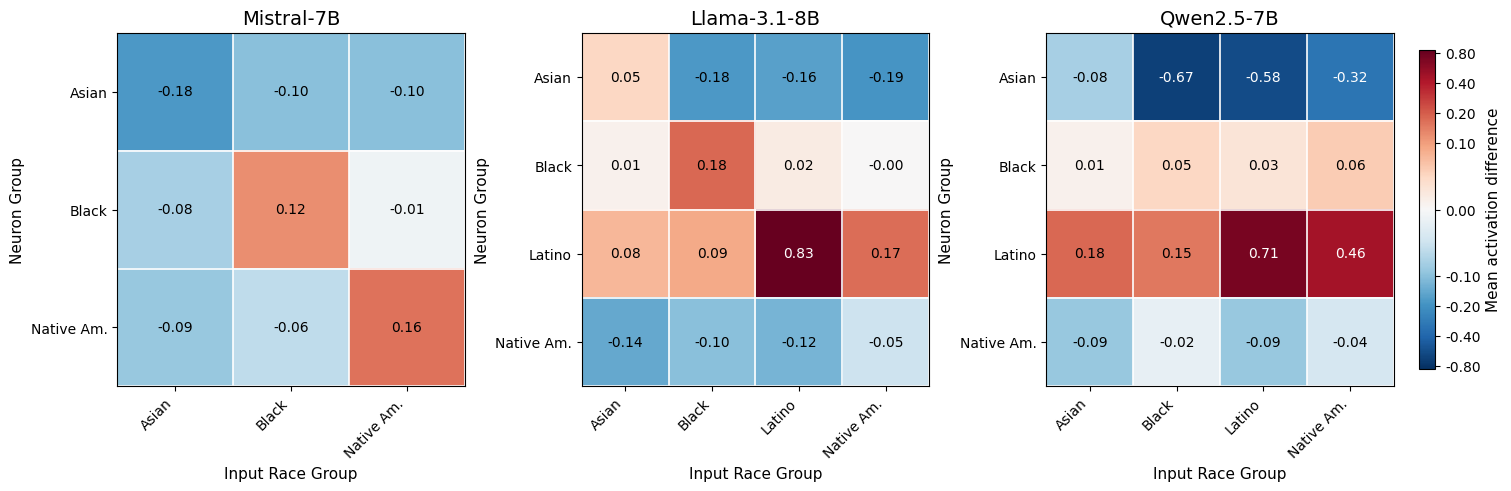}
\vspace{-.22in}
\caption{\small Mean activation values of race-associated neurons when processing text from each racial group (ToxiGen). Diagonal cells represent neurons processing their target group. Higher values (red) indicate stronger activation; lower values (blue) indicate weak, negative activations.}
\vspace{-1.5em}
\label{fig:activation_heatmaps}
\end{figure*}

\noindent \textbf{Neuron Activation Analysis.}
To examine whether the identified neurons respond more strongly to their associated demographic cues, we measure mean activation values when processing test samples from each group. Figure~\ref{fig:activation_heatmaps} displays these activation patterns as heatmaps, where diagonal cells correspond to neurons evaluated on samples from their associated group. The results show a partial diagonal pattern rather than uniform selectivity across all groups and models. Neurons associated with Latino cues show the clearest activation aligned with the target group in Llama-3.1-8B and Qwen2.5-7B, with diagonal values of 0.83 and 0.71, respectively. Neurons associated with Black cues also show positive diagonal activation in Mistral-7B and Llama-3.1-8B, although the pattern is weaker in Qwen2.5-7B. In contrast, neurons associated with Asian and Native American cues exhibit weaker and less consistent selectivity, with small or negative diagonal values in several models. Overall, these results suggest that neurons selected by probe direction alignment often capture cues associated with particular groups, but the strength of this association varies across groups and models.


\begin{table}[tb]
\centering
\small
\resizebox{\columnwidth}{!}{%
\begin{tabular}{llll}
\toprule
\textbf{Model} & \textbf{Group} & \textbf{Direct} & \textbf{Indirect} \\
\midrule
\multirow{3}{*}{Qwen2.5-7B}
    & White & 白 & Russian, 俄罗斯, Russia \\
    \cmidrule(lr){2-4}
    & Asian & Asian, Asia, Asians, 亚洲  & Chinese, Xia, Tibetan, China \\
    \cmidrule(lr){2-4}
    & Black/AA & African, 非洲, Africa, black & Hait, Haiti, Tropical, 热带  \\
\midrule
\multirow{3}{*}{Mistral-7B}
    & White &  & Moscow, Russian, Ukrain, Polish \\
    \cmidrule(lr){2-4}
    & Asian & Asian, Taiwan, Japanese, Malays & Korea, Korean, Asian, Vietnam \\
    \cmidrule(lr){2-4}
    & Black/AA & African, blacks, Negro, slavery & Caribbean, Cuba, Nigeria, Brazil \\
\midrule
\multirow{3}{*}{Llama-3.1-8B}
    & White &  & Russia, Kremlin, Putin, Moscow \\
    \cmidrule(lr){2-4}
    & Asian & Asian, Indonesian, Asia, Taiwanese & Cambodia, Chinese, wang, Buddhism \\
    \cmidrule(lr){2-4}
    & Black/AA & black, African, Afro, negro & Haiti, Caribbean, Dominican, Bahamas \\
\bottomrule
\end{tabular}%
}
\vspace{-.13in}
\caption{\small Comparison of top tokens from direct (race/ethnicity) vs. indirect (language/country) mentions in C-REACT. \textbf{WARNING: Some tokens reflect harmful stereotypes.} Full results in Appendix~\ref{app:creact_tokens}.
{\scriptsize Translations: 白 (white), 俄罗斯 (Russia), 非洲 (Africa), 热带 (tropical).}}
\vspace{-1.5em}
\label{tab:creact_tokens}
\end{table}
\subsection{C-REACT}
We next use C-REACT to compare how explicit demographic labels and indirect proxy cues, such as language or country, are represented internally. We train separate probes on direct and indirect mentions to evaluate whether each type captures distinct representations. Direct-mention probes achieve higher accuracy and comparable F1 to indirect probes (Appendix~\ref{app:probe_metrics}). This likely reflects the fact that direct cues are explicit and indirect data are sparser. Table~\ref{tab:creact_tokens} compares tokens projected by each probe type. Direct and indirect probes emphasize different semantic facets of race and ethnicity cues, while still showing partial overlap in the neurons they identify. Direct probes recover general racial and ethnic terminology (\textit{Asian, African, black, 白}), while indirect probes recover specific countries and regions associated with each group. For instance, the White indirect probe projects strongly to Russia and Eastern European terms across all models, reflecting the dataset composition where Russian is the most frequent language among White patients. Similarly, Black or African American indirect probes recover Caribbean and African nations (\textit{Haiti, Caribbean, Nigeria}). This pattern suggests that demographic labels and proxy cues are associated with partially overlapping but semantically different internal pathways, including explicit group labels and associated geographic or linguistic attributes.

\begin{table}[H]
\centering
\resizebox{\columnwidth}{!}{%
\begin{tabular}{llll}
\toprule
\textbf{Model} & \textbf{Group} & \textbf{Neuron} & \textbf{Top tokens} \\
\midrule
\multirow{6}{*}{Qwen2.5-7B}
    & \multirow{2}{*}{White} 
        & MLP.v$^{28}_{16880}$ & 英国, Dutch, French, Italian \\
    & & MLP.v$^{27}_{17660}$ & German, Germany, EU, euro \\
    \cmidrule(lr){2-4}
    & \multirow{2}{*}{Asian} 
        & MLP.v$^{27}_{6943}$ & Asian, Asia, Chinese, Indian \\
    & & MLP.v$^{27}_{217}$ & Asian, 亚洲, Asia, アジア\\
    \cmidrule(lr){2-4}
    & \multirow{2}{*}{Black/AA} 
        & MLP.v$^{28}_{11088}$ & racist, racism, Harlem, segregation \\
    & & MLP.v$^{27}_{2240}$ & black, 黑, Black, 黑色 \\
\midrule
\multirow{6}{*}{Mistral-7B}
    & \multirow{2}{*}{White} 
        & MLP.v$^{32}_{1606}$ & England, France, Europe, Switzerland \\
    & & MLP.v$^{32}_{9831}$ & European, Europe, EU, Euro \\
    \cmidrule(lr){2-4}
    & \multirow{2}{*}{Asian} 
        & MLP.v$^{32}_{4453}$ & Japanese, Korean, Japan, Taiwan \\
    & & MLP.v$^{31}_{2346}$ & Japanese, anime, Japan, Tokyo\\
    \cmidrule(lr){2-4}
    & \multirow{2}{*}{Black/AA} 
        & MLP.v$^{32}_{5923}$ & Black, 黑, Negro, African \\
    & & MLP.v$^{31}_{8715}$ & African, Africa, Kenya, Nigeria \\
\midrule
\multirow{5}{*}{Llama-3.1-8B}
    & \multirow{1}{*}{White} 
        & MLP.v$^{31}_{9094}$ & White, white, WHITE, 白 \\
    \cmidrule(lr){2-4}
    & \multirow{2}{*}{Asian} 
        & MLP.v$^{32}_{5691}$ & Chinese, China, Beijing, Shanghai \\
    & & MLP.v$^{29}_{5272}$ & Asia, Asia, continent, 亚洲\\
    \cmidrule(lr){2-4}
    & \multirow{2}{*}{Black/AA} 
        & MLP.v$^{30}_{7195}$ & Mississippi, Jamaica, Caribbean, Louisiana \\
    & & MLP.v$^{29}_{13826}$ & African, african, Afro, negro \\
\bottomrule
\end{tabular}%
}
\vspace{-.1in}
\caption{\small Top race-associated neurons from C-REACT \textbf{direct} mentions (explicit race/ethnicity). \textbf{WARNING: Some tokens reflect harmful stereotypes.} Full results in Appendix~\ref{app:creact_direct_neurons}.
{\scriptsize Translations: 英国 (England), 亚洲 (Asia), アジア (Asia), 黑 (Black), 黑色 (Black color), 白 (White).}}
\vspace{-1em}
\label{tab:creact_direct_neurons}
\end{table}

Tables~\ref{tab:creact_direct_neurons} and~\ref{tab:creact_indirect_neurons} present race-associated neurons identified from direct and indirect probes respectively. The neuron projections mirror the probe token patterns: direct mention neurons recover explicit demographic terms (\textit{Asian, Black, African, 白(White)}), while indirect mention neurons recover geographic and cultural associations (\textit{Russia, Moscow, Vietnam, Caribbean}). As in ToxiGen, we observe neurons whose projected tokens reflect harmful associations. Qwen2.5-7B's $\text{MLP.v}^{28}_{11088}$ projects to \textit{racist, racism, Harlem, segregation}, and several Black/AA neurons project to terms related to slavery. The persistence of such projected associations across both general and clinical domains indicates that harmful stereotype-related associations are embedded within these models and are not limited to specific task contexts.

\vspace{-0.5em}
\begin{table}[H]
\centering
\resizebox{\columnwidth}{!}{%
\begin{tabular}{llll}
\toprule
\textbf{Model} & \textbf{Group} & \textbf{Neuron} & \textbf{Top tokens} \\
\midrule
\multirow{6}{*}{Qwen2.5-7B}
    & \multirow{2}{*}{White} 
        & MLP.v$^{27}_{17660}$ & German, Germany, 荷兰, euro \\
    & & MLP.v$^{26}_{3382}$ & Russians, Russia, 俄罗斯, Moscow \\
    \cmidrule(lr){2-4}
    & \multirow{2}{*}{Asian} 
        & MLP.v$^{28}_{13406}$ & Japanese, 日本, Tokyo, Osaka \\
    & & MLP.v$^{25}_{2001}$ & Vietnam, Viet, Nguyen, Vietnamese\\
    \cmidrule(lr){2-4}
    & \multirow{2}{*}{Black/AA} 
        & MLP.v$^{25}_{10230}$ & 非洲, African, slave, slavery \\
    & & MLP.v$^{25}_{10739}$ & African, Africa, Ghana, Nigerian \\
\midrule
\multirow{3}{*}{Mistral-7B}
    & \multirow{2}{*}{White} 
        & MLP.v$^{32}_{2399}$ & Russian, Vlad, Moscow, Soviet \\
    & & MLP.v$^{29}_{260}$ & Italian, Italy, Giovanni, Francesco \\
    \cmidrule(lr){2-4}
    & \multirow{1}{*}{Black/AA} 
        & MLP.v$^{31}_{8715}$ & African, Africa, Kenya, Nigeria \\
\midrule
\multirow{6}{*}{Llama-3.1-8B}
    & \multirow{2}{*}{White} 
        & MLP.v$^{32}_{10606}$ & Russian, Moscow, Soviet, Putin \\
    & & MLP.v$^{29}_{4193}$ & Czech, Hungarian, Slovak, Budapest\\
    \cmidrule(lr){2-4}
    & \multirow{2}{*}{Asian} 
        & MLP.v$^{32}_{5691}$ & Chinese, China, Beijing, Shanghai \\
    & & MLP.v$^{29}_{10616}$ & Indian, India, Bollywood, Mumbai\\
    \cmidrule(lr){2-4}
    & \multirow{2}{*}{Black/AA} 
        & MLP.v$^{29}_{6824}$ & tropical, jungle, Congo, Caribbean \\
    & & MLP.v$^{29}_{13826}$ & African, african, Afro, negro \\
\bottomrule
\end{tabular}%
}
\vspace{-.1in}
\caption{\small Top race-associated neurons from C-REACT \textbf{indirect} mentions (language/country). \textbf{WARNING: Some tokens reflect harmful stereotypes.} Full results in Appendix~\ref{app:creact_indirect_neurons}. 
{\scriptsize Translations: 荷兰 (Netherlands), 俄罗斯 (Russia), 日本 (Japan), 非洲 (Africa).}}
\vspace{-1em}
\label{tab:creact_indirect_neurons}
\end{table}

\subsection{Neuron Intervention}
Finally, we use C-REACT indirect mentions to test whether the neuron groups identified above are causally involved in the dominant misclassification patterns. Using a template prompt shown in Figure~\ref{fig:prompt}, we prompt each model to predict patient race based on clinical text containing only indirect cues such as language or country information, and then manipulate the identified neurons to measure changes in biased prediction patterns. 
\begin{figure}[!ht]
\small
\begin{mdframed}[linewidth=0.2pt, innerleftmargin=4pt, innerrightmargin=4pt, innertopmargin=4pt, innerbottommargin=4pt]
\raggedright
\texttt{Please read the following sentence from a clinical note. Based on the information contained within the sentence, determine which of the following races or ethnicities the patient belongs to. Please respond with only one option.}\\[0.2em]
\texttt{Options: [White, Black or African American, Asian]}\\[0.2em]
\texttt{Clinical Note: "\{text\}"}\\[0.2em]
\texttt{Inferred Race or Ethnicity:}
\end{mdframed}
\vspace{-.1in}
\caption{\small Prompt template for race prediction on C-REACT indirect mentions.}
\vspace{-1.5em}
\label{fig:prompt}
\end{figure}

\noindent \textbf{Baseline Classification.}
Table~\ref{tab:error_patterns} shows misclassification patterns across the three models. The dominant error type varies by model: for Qwen2.5-7B and Llama-3.1-8B, White→Asian misclassification is the primary error, accounting for 75.0\% and 95.6\% of errors respectively. Llama's bias is the most pronounced, with 395 of 537 White patients incorrectly classified as Asian. In contrast, Mistral-7B shows a different pattern: its dominant error is White→Black/AA (76.2\% of errors). This divergence suggests that models rely on different demographic cues during inference.

\begin{table}[H]
\centering
\resizebox{\columnwidth}{!}{%
\begin{tabular}{lccc}
\toprule
\textbf{Error Type} & \textbf{Qwen2.5-7B} & \textbf{Mistral-7B} & \textbf{Llama-3.1-8B} \\
\midrule
White→Asian & \textbf{27} & 4 & \textbf{395} \\
White→Black/AA & 4 & \textbf{16} & 5 \\
Black/AA→White & 4 & 0 & 0 \\
Black/AA→Asian & 0 & 0 & 10 \\
Asian→White & 1 & 1 & 1 \\
Asian→Black/AA & 0 & 0 & 2 \\
\midrule
Total Errors & 36 & 21 & 413 \\
\textbf{Dominant Error \%} & \textbf{75.0\%} & \textbf{76.2\%} & \textbf{95.6\%} \\
\bottomrule
\end{tabular}
}
\vspace{-.1in}
\caption{\small Misclassification patterns on C-REACT indirect mentions. The dominant error type (bold) varies across models: White→Asian for Qwen and Llama, White→Black/AA for Mistral.}
\label{tab:error_patterns}
\vspace{-0.6cm}
\end{table}

\vspace{.1in}
\noindent \textbf{Activation Patterns.} To investigate what drives these biases, we measure activation levels for all neuron groups across all classification outcomes (Table~\ref{tab:activation_full}). We observe a strong correspondence between neuron groups exhibiting consistently high activation and dominant error directions identified in Table~\ref{tab:error_patterns}. 
For Qwen2.5-7B, which primarily misclassifies White patients as Asian, the Asian Direct neurons show consistently high positive activation regardless of ground truth or prediction. Similarly, Mistral-7B's tendency toward White $\rightarrow$ Black/AA errors aligns with elevated activity in Black/AA Direct neurons across most scenarios. Llama-3.1-8B presents a different pattern: while its dominant error is also White $\rightarrow$ Asian, Asian Indirect neurons show consistently high activation across scenarios.
These patterns reveal that neuron groups with consistently high activation often correspond to the dominant misclassification directions, suggesting they may play a causal role in bias.

\begin{table*}[!ht]
\centering
\small
\resizebox{\textwidth}{!}{%
\begin{tabular}{ll|ccccccccc}
\toprule
& & \multicolumn{9}{c}{\textbf{Classification Outcome (Actual → Predicted)}} \\
\textbf{Model} & \textbf{Neuron Group} & W→W & W→B & W→A & B→W & B→B & B→A & A→W & A→B & A→A \\
\midrule
\multirow{6}{*}{Qwen2.5-7B}
& Asian Direct & \textbf{+6.04} & \textbf{+6.25} & \textbf{+5.85} & \textbf{+3.50} & \textbf{+5.71} & --- & \textbf{+4.66} & --- & \textbf{+3.86} \\
& Asian Indirect & $-$1.03 & $-$1.02 & $-$0.96 & +0.31 & $-$0.06 & --- & $-$1.55 & --- & $-$0.33 \\
& Black/AA Direct & $-$0.74 & $-$0.04 & $-$0.40 & +0.28 & +0.83 & --- & +0.27 & --- & +0.27 \\
& Black/AA Indirect & $-$0.57 & +0.40 & $-$0.63 & +5.80 & +3.30 & --- & $-$0.39 & --- & $-$0.78 \\
& White Direct & $-$3.25 & $-$3.49 & $-$3.80 & $-$2.06 & $-$3.05 & --- & $-$4.87 & --- & $-$4.01 \\
& White Indirect & +0.36 & +0.31 & $-$0.20 & $-$0.90 & $-$1.24 & --- & $-$0.33 & --- & $-$1.31 \\
\midrule
\multirow{5}{*}{Mistral-7B}
& Asian Direct & +0.22 & +0.24 & +0.14 & --- & $-$0.15 & --- & +0.36 & --- & +0.19 \\
& Black/AA Direct & \textbf{+0.82} & \textbf{+0.89} & \textbf{+0.64} & --- & $-$0.07 & --- & \textbf{+1.00} & --- & \textbf{+0.41} \\
& Black/AA Indirect & $-$0.56 & $-$0.42 & $-$0.55 & --- & +0.19 & --- & $-$0.52 & --- & $-$0.44 \\
& White Direct & +0.10 & $-$0.06 & $-$0.02 & --- & $-$0.47 & --- & +0.33 & --- & +0.37 \\
& White Indirect & $-$0.01 & +0.24 & $-$0.17 & --- & $-$0.27 & --- & $-$0.44 & --- & $-$0.32 \\
\midrule
\multirow{6}{*}{Llama-3.1-8B}
& Asian Direct & $-$0.53 & $-$0.59 & $-$0.59 & --- & $-$0.57 & $-$0.54 & $-$0.41 & $-$0.51 & $-$0.55 \\
& Asian Indirect & \textbf{+0.27} & \textbf{+0.28} & \textbf{+0.33} & --- & \textbf{+0.36} & \textbf{+0.43} & \textbf{+0.29} & \textbf{+0.34} & \textbf{+0.47} \\
& Black/AA Direct & $-$1.07 & $-$1.43 & $-$1.20 & --- & $-$1.97 & $-$1.66 & $-$0.54 & $-$1.60 & $-$0.96 \\
& Black/AA Indirect & $-$0.01 & +0.01 & $-$0.00 & --- & +0.04 & +0.07 & +0.03 & +0.03 & +0.03 \\
& White Direct & +0.11 & +0.37 & +0.24 & --- & +0.51 & +0.49 & +0.62 & +0.34 & +0.70 \\
& White Indirect & +0.22 & +0.28 & +0.32 & --- & +0.01 & +0.04 & $-$0.04 & +0.01 & $-$0.01 \\
\bottomrule
\end{tabular}%
}
\vspace{-.1in}
\caption{\small Mean neuron activation scores on C-REACT indirect mention prompts, grouped by classification outcome (\textit{actual} $\rightarrow$ \textit{predicted}). \textbf{W/B/A} denote \textbf{White / Black/AA / Asian} (e.g., \textbf{W$\rightarrow$B}: actual White, predicted Black/AA; \textbf{W$\rightarrow$W}: correct). Positive values indicate neuron group writes in direction of its output vectors (strengthening the associated signal during generation), while negative values indicate writing in the opposite direction (weakening it). ``---'' indicates no samples for that outcome.}
\label{tab:activation_full}
\vspace{-1.5em}
\end{table*}

\begin{figure}[ht]
\centering
\includegraphics[width=\columnwidth]{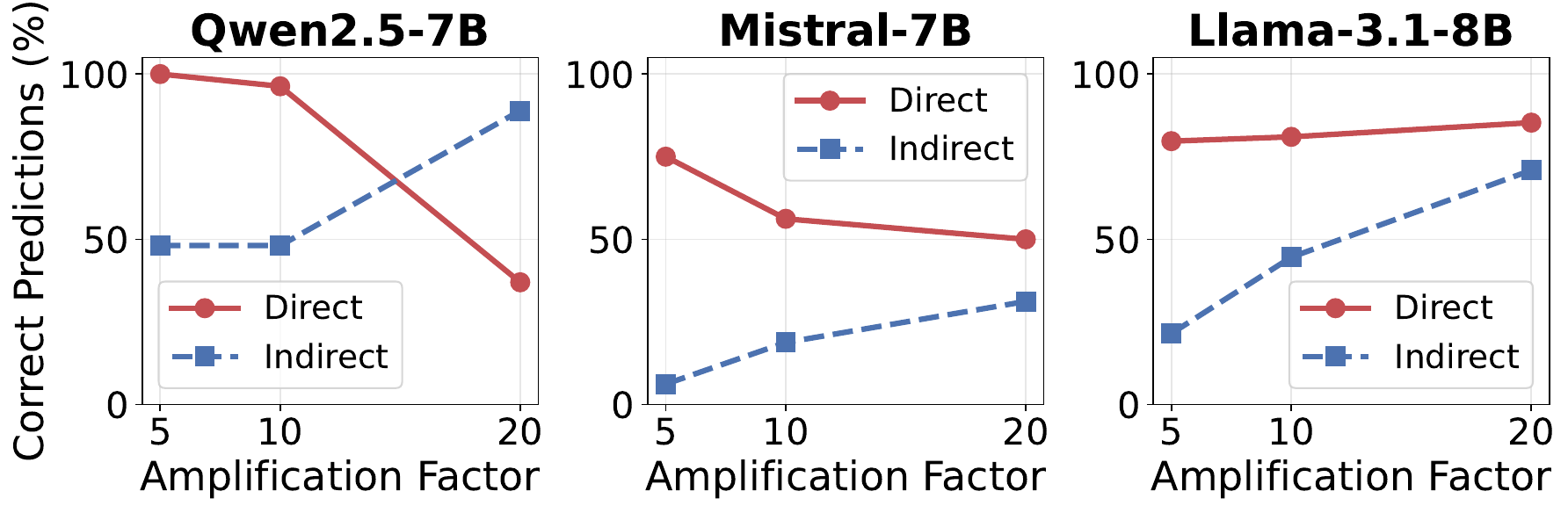}
\vspace{-.3in}
\caption{\small Correct prediction rates after neuron intervention across amplification factors. Direct neuron intervention (solid) generally outperforms Indirect intervention (dashed), demonstrating that neurons associated with explicit racial terminology have stronger causal influence on predictions.}
\vspace{-1em}
\label{fig:intervention_line}
\end{figure}

\begin{figure}[ht]
\centering
\includegraphics[width=\columnwidth]{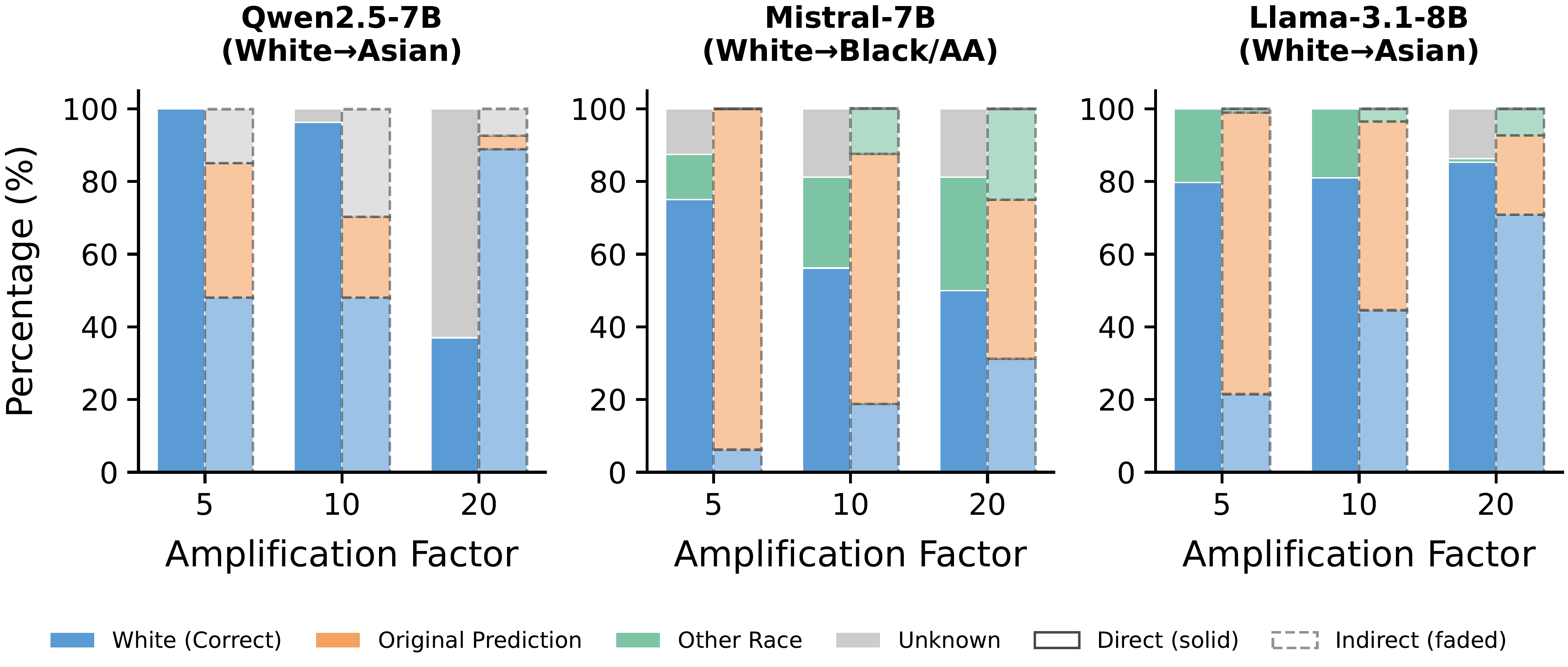}
\vspace{-.25in}

\caption{\small Prediction distribution after neuron intervention on misclassified samples. Direct intervention (solid bars) removes the original dominant error label (orange `Original Prediction' bars = 0\%) across all models and factors, while Indirect intervention (faded bars) leaves residual bias. Higher amplification factors increase Unknown responses (gray)}
\vspace{-0.5em}
\label{fig:intervention_stacked}
\end{figure}

\noindent \textbf{Intervention Results.}
Having identified neuron groups associated with dominant error patterns, we test whether intervening on these neurons changes the model's predictions. Specifically, we evaluate the intervention using three amplification factors (5, 10, 20) to assess the effect of intervention strength. Figure~\ref{fig:intervention_line} compares correct prediction rates between Direct and Indirect neuron intervention, while Figure~\ref{fig:intervention_stacked} shows the full prediction distribution across all conditions.

\noindent \textbf{Direct vs.\ Indirect Neurons.}
Across all three models, Direct neuron intervention demonstrates stronger causal efficacy than Indirect intervention (Figure~\ref{fig:intervention_line}). At factor 5, Direct intervention achieves substantially higher correct prediction rates across all models compared to Indirect intervention. More importantly, Direct intervention eliminates the original dominant error label in this targeted setting across all models and amplification factors (Figure~\ref{fig:intervention_stacked}), while Indirect intervention leaves residual bias. This gap is also consistent with Llama-3.1-8B's pattern: although the Asian Indirect group has higher mean activation, higher activation does not necessarily mean stronger causal influence on final prediction. Indirect cues tend to reflect broad, proxy signals that can be supported by multiple parts of the network, so steering one indirect group may be partly compensated elsewhere and leads to a smaller behavioral change. By contrast, Direct neurons are more directly tied to producing explicit race labels, which makes intervening on them more effective.

\noindent \textbf{Amplification Factor Selection.}
The choice of amplification factor involves a tradeoff between targeted prediction correction and model stability. We tested factors of 5, 10, and 20, representing increasingly aggressive intervention.
Across all models, factor 5 yields the best balance. Qwen2.5-7B achieves 100\% correct predictions with no Unknown outputs at factor 5, but destabilizes at factor 20 (63\% Unknown responses). Mistral-7B reaches 75\% correct predictions at factor 5, with higher factors increasingly shifting predictions toward Asian rather than the correct White label. Llama-3.1-8B performs similarly at factors 5 and 10 (around 80\% correct), with factor 20 introducing Unknown responses. These results suggest moderate intervention strength suffices to alter predictions via race-associated neurons; we adopt factor 5 as the default.

\section{Discussion and Conclusion} 
Our results indicate that the way race and ethnicity cues are reflected in LLMs is central to understanding how demographic bias emerges across tasks. The first important finding is that \textbf{racial and ethnic concepts are distributed across many internal units rather than localized to a small set of neurons}. Importantly, this distribution is not arbitrary: models decompose race and ethnicity cues into multiple, interpretable semantic facets, such as explicit group labels and associated geographic or linguistic attributes. Across both ToxiGen and C-REACT, these facets appear as distinct internal representations rather than single abstract concepts (Tables~\ref{tab:probe_tokens},~\ref{tab:neurons},~\ref{tab:creact_tokens}). Notably, stereotype-related and historically harmful associations are present across models despite differences in training data and geographic origin, suggesting that bias mitigation cannot rely on a universal map of demographic features but requires model-specific localization. 

Secondly, due to this representational structure, the same internal components can be reused across different task contexts, sometimes in ways that lead to biased behavior. \textbf{Neurons associated with racial concepts are present in all three models, yet their influence on predictions varies substantially depending on whether the input associates strongly with the proxy cues.} The same representations that capture demographic information in neutral contexts can lead to biased predictions when activated in contexts where race is irrelevant.

\textit{Crucially, the presence of such representations is not inherently problematic.} Rather, bias arises from how these representations are operationalized during inference. Our intervention \textit{did not erase} racial knowledge from the models; instead, it modulated how this knowledge was reused in task-specific settings. This distinction is critical: pretrained representations reflect what models learn about the world, whereas task-dependent bias reflects when and how those representations are inappropriately applied. 


To summarize, we provide a mechanistic analysis of internal features associated with race and ethnicity cues in LLMs. These findings suggest that mitigating demographic bias requires examining internal feature structure and task-specific reuse, beyond outcome-level intervention.

\section*{Ethical Statement}
This work examines how large language models respond to and internally represent textual cues associated with race and ethnicity, which necessarily involves reporting sensitive content including stereotypes and historically offensive terminology. We present these findings to expose potential bias, not to amplify it. We acknowledge that racial categories are socially constructed and vary across cultures; our use of race categories reflects the structure of the datasets rather than an endorsement of these taxonomies. We use all publicly available datasets and models in accordance with their released licenses and terms of use.
\section*{Limitations}
Our study has several limitations. First, our clinical analysis is limited to three racial groups (White, Asian, and Black/AA) due to data availability; C-REACT is the only suitable dataset we identified for this task, and other racial categories lacked sufficient representation for reliable probe training. This restricts our ability to assess whether the associated patterns we observe generalize to other demographic groups.

Second, our neuron identification relies on cosine similarity between MLP value vectors and probe directions, which captures neurons with strong linear alignment to race representations. However, racial representations may also be distributed across neurons that contribute through more complex, nonlinear interactions not detected by our method. The observation that high activation does not always imply causal influence (e.g., Llama's Asian Indirect neurons) suggests that our identification approach captures a subset of race-associated neurons but may miss components that operate through alternative pathways.

Finally, our intervention experiments target neurons identified from a single task (race prediction from clinical text), and we do not evaluate whether the same neurons drive biased behavior in downstream clinical applications such as diagnosis prediction or treatment recommendation. Future work should examine whether interventions transfer across tasks or require task-specific neuron identification.

\bibliography{custom}

\clearpage
\appendix
\section{Appendix}
\label{sec:appendix}
\subsection{Probe Performance}
\label{app:probe_metrics}

\begin{table}[H]
\centering
\resizebox{\columnwidth}{!}{%
\begin{tabular}{lcc}
\toprule
Model & Acc. (\%) & Macro-F1 \\
\midrule
Qwen2.5-7B-IT   & 74.88 & 0.74 \\
Llama-3.1-8B-IT & 77.98 & 0.77 \\
Mistral-7B-IT   & 75.20 & 0.74 \\
\bottomrule
\end{tabular}
}
\caption{Test set performance of multi-class linear probes trained on \textbf{ToxiGen} (4-way: asian/black/latino/native\_american). We report Accuracy and Macro-F1.}
\label{tab:probe_toxigen}
\end{table}

\begin{table}[H]
\centering
\small
\resizebox{\columnwidth}{!}{%
\begin{tabular}{lcccc}
\toprule
& \multicolumn{2}{c}{Direct} & \multicolumn{2}{c}{Indirect} \\
\cmidrule(lr){2-3}\cmidrule(lr){4-5}
Model & Acc. (\%) & Macro-F1 & Acc. (\%) & Macro-F1 \\
\midrule
Qwen2.5-7B-IT   & 90.36 & 0.79 & 81.61 & 0.79   \\
Llama-3.1-8B-IT & 93.98 & 0.85 & 80.46 & 0.80 \\
Mistral-7B-IT   & 90.66 & 0.70 & 75.86 & 0.74 \\
\bottomrule
\end{tabular}
}
\caption{Test set performance of linear probes trained on \textbf{C-REACT} (3-way: White/Black/AA/Asian), using Direct vs.\ Indirect prompt variants.}
\label{tab:probe_creact}
\end{table}

\subsection{Probe Token Projections (ToxiGen)}
\label{app:ToxiGen_tokens}
Table~\ref{tab:appendix_toxigen_token} presents the complete top-20 tokens projected by each race direction probe for all three models.

\begin{table*}[ht]
\centering
\resizebox{\textwidth}{!}{%
\begin{tabular}{lll}
\toprule
\textbf{Model} & \textbf{Group} & \textbf{Top 20 Tokens} \\
\midrule
\multirow{5}{*}{Qwen2.5-7B}
& Asian & Asian, 亚洲, Asian, Chun, Chinese, CJK, Yuan, 东亚, {\thaifont ่ว}, chinese, 霹, 恶, Chinese, ActivityCreated, chner, Chu, ObjectContext, lobals, Hong, chin \\
\cmidrule(lr){2-3}
& Black & ="\{\}/>, 相应, orsk, 台阶, 截, {\cyrfont либо}, 没有想到, Rudd, 钮, .BLL, .GetDirectoryName, ]interface, setError, 鹕, Beaut, 夙, :async, 翾, 绶, SingleOrDefault \\
\cmidrule(lr){2-3}
& Latino & nesty, Mex, .Span, .Shared, 垭, ."/, /items, SetLastError, ucher, getTime, .onclick, enticate, ERVE, mex, .Stretch, mexico, tü, evenodd, asher, 年之久 \\
\cmidrule(lr){2-3}
& Native Am. & natives, native, .Native, native, Native, Native, \_native, indigenous, coma, .native, -native, ""\}, 達, ings, NAV, ITERAL, reservation, INGS, 킴, 葆 \\
\midrule
\multirow{5}{*}{Mistral-7B}
& Asian & Chinese, Asian, China, Korean, Taiwan, Hong, inese, Japanese, ', Shanghai, Beijing, Asia, omi, {\thaifont ่}, ß, bt, Roose, MMMM, omy, agh \\
\cmidrule(lr){2-3}
& Black & /******/, corpor, Coff, Email, black, uh, African, publicly, spell, ta, white, email, black, Black, Palm, Parl, riel, Black, mask, rele \\
\cmidrule(lr){2-3}
& Latino & Mexico, Salvador, jd, Colombia, Chile, ¡, Colomb, ass, Mexican, Skip, ranch, Mex, Santiago, ully, Argent, jal, Partido, sketch, skip, aca \\
\cmidrule(lr){2-3}
& Native Am. & Indians, trib, Native, tribes, Indian, medicine, Native, ye, Medicine, AD, ingle, minister, ilo, Inga, tribe, iga, unda, Trib, uti, erset \\
\midrule
\multirow{5}{*}{Llama-3.1-8B}
& Asian & Asian, Asian, Mandarin, CJK, erse, ležit, ibold, Ding, inese, asian, rysler, china, ern, Chinese, Asia, ihan, Asians, ertainment, ź, Euras \\
\cmidrule(lr){2-3}
& Black & CELER, Black, isay, Black, \_black, urgeon, -black, adeon, arp, cie, anja, .black, 黑, {\arabfont ورا}, -hit, /black, {\arabfont جل}, ouse, {\arabfont بلند}, {\arabfont وتر} \\
\cmidrule(lr){2-3}
& Latino & ucwords, 妙, udos, Buchanan, .si, '', oppos, aside, i, City, rip, Mundo, odí, \_BORDER, mant, cloak, \_gid, .ul, iyon, ivery \\
\cmidrule(lr){2-3}
& Native Am. & Native, Native, native, natives, native, -native, Indians, \_Native, \_native, indigenous, .native, /native, Indigenous, tribes, RIPT, Indian, reservation, .Native, ative, Reservation \\
\bottomrule
\end{tabular}
}
\caption{Full top-20 probe token projections for all models (ToxiGen). Tokens are listed in descending order of projection score.
{\scriptsize Translations: 亚洲 (Asia), 东亚 (East Asia), {\thaifont ่ว} (Thai mark), 霹 (thunderclap), 恶 (evil), 相应 (corresponding), 台阶 (steps), 截 (cut), {\cyrfont либо} (either), 没有想到 (didn’t expect), 钮 (button), 鹕 (pelican), 夙 (early), 翾 (swift flight), 绶 (ribbon), 垭 (mountain pass), tü (door), 年之久 (for years), 達 (reach), {\greekfont 킴} (Kim), 葆 (preserve), ß (ss), 黑 (black), {\arabfont ورا} (behind), {\arabfont جل} (skin), {\arabfont بلند} (tall), {\arabfont وتر} (string), 妙 (wonderful), {\cyrfont сов} (owl), {\cyrfont станов} (become), ź (z), šem (name), odí (I hated).}
}
\label{tab:appendix_toxigen_token}
\end{table*}

\subsection{Race-Associated Neurons (ToxiGen)}
\label{app:ToxiGen_neurons}
From Table~\ref{tab:appendix_toxigen_qwen} to Table~\ref{tab:appendix_toxigen_llama} present the complete lists of race-associated neurons identified from ToxiGen for the three models.

\begin{table*}[h!]
\centering
\resizebox{\textwidth}{!}{%
\begin{tabular}{llll}
\toprule
\textbf{Model} & \textbf{Group} & \textbf{Neuron} & \textbf{Top Tokens} \\
\midrule
\multirow{26}{*}{Qwen2.5-7B}
 & \multirow{5}{*}{Asian} & MLP.v$_{13406}^{28}$ & 在日本, Japanese, 日本, Japanese, Japan, Japan, 日本人, Tokyo, japan, 东京, japan, 일본, 日军, japanese, japon, Nhật, 日本の, Tok, Hiro, jap, Osaka \\
 \cmidrule{3-4}
 & & MLP.v$_{5983}^{28}$ & Chi, chi, Chin, Chi, chin, chi, \_chi, Xi, Hu, xi, Hu, xi, 人民币, Xi, chin, Huawei, 华为, 华夏, (xi, hu \\
 \cmidrule{3-4}
 & & MLP.v$_{8641}^{27}$ & ch, 是中国, Ch, 为中国, Chinese, 成为中国, 由中国, 与中国, 对中国, China, china, \_ch, 在中国, China, Ch, china, Chinese, -Ch, CH, chin \\
 \cmidrule{3-4}
 & & MLP.v$_{217}^{27}$ & Asian, 亚洲, Asia, Asian, Asians, Asia, asian, asia, アジア, 亚, 亞, Asi, asia, asiat, asi, 亚太, {\thaifont เอเช}, \_As, asi, AS \\
 \cmidrule{3-4}
 & & MLP.v$_{15029}^{25}$ & Chinese, China, China, Chinese, chinese, china, china, 中国, 中国的, -China, 중국, {\arabfont الصين}, 中國, Asian, 在中国, 是中国, 亚洲, 由中国, Asian, Asia \\
\cmidrule(lr){2-4}
 & Black & MLP.v$_{2240}^{27}$ & black, 黑, black, Black, Black, 黑色, -black, BLACK, BLACK, \_black, blacks, .black, /black, , ブラック, .Black, ⿊, 黑白, \_BLACK, blacklist \\
\cmidrule(lr){2-4}
 & \multirow{3}{*}{Latino} & MLP.v$_{4781}^{28}$ & Latin, Latin, 拉丁, latin, latin, LATIN, latino, 巴西, Latino, LAT, Latina, latina, Lat, Brazil, \_LAT, 阿根廷, Lat, Brazil, lat, LAT \\
 \cmidrule{3-4}
 & & MLP.v$_{9876}^{28}$ & Spanish, 西班牙, 湘, Portuguese, Hispanic, Spanish, 贵州, Brazilian, ¡, ¡, 贵州省, 黔, 葡萄牙, Juan, Chile, spanish, 贵阳, Brazil, Juan, Spain \\
 \cmidrule{3-4}
 & & MLP.v$_{18125}^{27}$ & Spanish, 西班牙, Hispanic, Spanish, Chile, Mexican, 贵州, 墨西哥, Spain, Brazilian, Juan, Mex, Juan, 巴西, 阿根廷, 贵州省, Mexico, 湘, spanish, Mexico \\
\cmidrule(lr){2-4}
 & \multirow{3}{*}{Native Am.} & MLP.v$_{3458}^{25}$ & native, native, 自然, natural, Native, Native, 本土, 天然, -native, indigenous, natural, natives, naturally, Natural, /native, nat, Natural, \_native, .native, 自发 \\
 \cmidrule{3-4}
 & & MLP.v$_{7087}^{25}$ & native, 故, native, Native, 故乡, 本土, 家乡, -native, Native, natives, hometown, .native, \_native, home, homeland, род, 祖国, /native, birth, quê \\
 \cmidrule{3-4}
 & & MLP.v$_{11197}^{25}$ & 殖民, colonial, colon, colony, colon, colonies, Colonial, Colon, Colon, Colony, imperial, Imperial, olon, icolon, -col, 帝国, ocol, colonization, dec, \_COL \\
\bottomrule
\end{tabular}
}
\caption{Full top-20 tokens for race-associated neurons in Qwen2.5-7B (ToxiGen).
{\scriptsize Translations: 在日本 (in Japan), 日本 (Japan), 日本の (Japan), 日本人 (Japanese), 东京 (Tokyo), 日军 (Japanese army), 是中国 (China), 为中国 (China), 成为中国 (China), 由中国 (China), 与中国 (China), 对中国 (China), 在中国 (China), 中国 (China), 中國 (China), 中国的 (China), 亚洲 (Asia), 亚 (Asia), 亞 (Asia), 亚太 (Asia-Pacific), 人民币 (Renminbi), 华为 (Huawei), 华夏 (China), 黑 (black), 黑色 (black), 黑白 (black-and-white), ⿊ (black), {\greekfont ブラック} (black), 拉丁 (Latin), 巴西 (Brazil), 阿根廷 (Argentina), 西班牙 (Spain), 墨西哥 (Mexico), 葡萄牙 (Portugal), 贵州 (Guizhou), 贵州省 (Guizhou), 贵阳 (Guiyang), 黔 (Guizhou), 湘 (Hunan), 自然 (natural), 天然 (natural), 本土 (native), 故乡 (hometown), 家乡 (hometown), 故 (hometown), 祖国 (motherland), 殖民 (colonial), 帝国 (empire), {\greekfont アジア} (Asia), {\greekfont 일본} (Japan), {\greekfont 중국} (China), {\thaifont เอเช} (Asia), {\arabfont الصين} (China), {\cyrfont ев} (Jew), {\cyrfont евр} (Jew), {\cyrfont род} (homeland), ¡ (!), Việt: Nhật (Japan), Português: quê (homeland).}}
\label{tab:appendix_toxigen_qwen}
\end{table*}

\begin{table*}[h!]
\centering
\resizebox{\textwidth}{!}{%
\begin{tabular}{llll}
\toprule
\textbf{Model} & \textbf{Group} & \textbf{Neuron} & \textbf{Top Tokens} \\
\midrule
\multirow{10}{*}{Mistral-7B}
 & Asian & MLP.v$_{4453}^{32}$ & Japanese, Korean, Kol, Japan, Kor, Kaz, Sak, Taiwan, Korea, Nak, Kur, Kom, Bangl, Pakistan, Kab, Kon, Pak, Tibet, Ku, Asian \\
\cmidrule(lr){2-4}
 & \multirow{3}{*}{Black} & MLP.v$_{5923}^{32}$ & Black, black, Black, black, blacks, 黑, 黒, Negro, BL, blk, African, BL, {\cyrfont чер}, Afr, Dark, {\cyrfont ч}, dark, {\cyrfont Чер}, {\cyrfont лек}, 블 \\
 \cmidrule{3-4}
 & & MLP.v$_{12572}^{30}$ & Black, black, Black, black, blacks, 黑, BL, 黒, BL, Negro, blk, {\cyrfont чер}, dark, African, Afr, 블, {\cyrfont лек}, dark, {\cyrfont Чер}, {\cyrfont ч} \\
 \cmidrule{3-4}
 & & MLP.v$_{13186}^{29}$ & black, black, Black, blacks, Black, white, 黑, white, whites, African, 黒, Negro, 白, Afr, White, White, BL, {\cyrfont чер}, Af, brown \\
\cmidrule(lr){2-4}
 & \multirow{2}{*}{Native Am.} & MLP.v$_{3440}^{31}$ & imperial, fasc, Imperial, colonial, militar, Kent, /******/, colon, popul, racist, gent, antal, neo, provinc, Emitter, colon, carriage, TES, omena, ounds \\
 \cmidrule{3-4}
 & & MLP.v$_{12205}^{29}$ & native, native, Native, Native, nat, ind, ab, igenous, Ma, nat, Ab, Ma, abor, Nat, primitive, Nav, born, Mas, nav, Ab \\
\bottomrule
\end{tabular}
}
\caption{Full top-20 tokens for race-associated neurons in Mistral-7B (ToxiGen).
{\scriptsize Translations: 黑 (black), 黒 (black), {\cyrfont чер} (black), {\cyrfont ч} (ch), {\cyrfont Чер} (black), {\cyrfont лек} (lek), 블 (black), 白 (white).}}
\label{tab:appendix_toxigen_mistral}
\end{table*}

\begin{table*}[h!]
\centering
\resizebox{\textwidth}{!}{%
\begin{tabular}{llll}
\toprule
\textbf{Model} & \textbf{Group} & \textbf{Neuron} & \textbf{Top Tokens} \\
\midrule
\multirow{14}{*}{Llama-3.1-8B}
 & \multirow{3}{*}{Asian} & MLP.v$_{5691}^{32}$ & Chinese, China, Chinese, China, chinese, china, -China, Beijing, 中国, Çin, 중국, Shanghai, 中國, china, {\arabfont چین}, 中国, Zhang, Jiang, Tencent, Guang \\
 \cmidrule{3-4}
 & & MLP.v$_{14299}^{31}$ & Li, yuan, Dong, Huang, Liu, Wang, Chen, Yang, Tian, Zhou, Ding, Wu, dong, dong, Feng, wang, Zhang, Qin, Jiang, Guang \\
 \cmidrule{3-4}
 & & MLP.v$_{12566}^{30}$ & East, East, EAST, Eastern, east, -East, eastern, 東, Eastern, east, 东, -east, 東, Doğu, 东, Đông, {\arabfont شرق}, vých, {\cyrfont вост}, orient \\
\cmidrule(lr){2-4}
 & \multirow{2}{*}{Black} & MLP.v$_{7195}^{30}$ & Mississippi, Jamaica, Jama, Caribbean, Louisiana, Trinidad, Haiti, LSU, Ghana, Hait, Baton, Harlem, Nigeria, negro, Negro, Jazz, Bahamas, Memphis, Nigerian, Zimbabwe \\
 \cmidrule{3-4}
 & & MLP.v$_{13826}^{29}$ & African, african, Africans, Africa, Africa, afr, Afro, Afr, frican, negro, Af, Afrika, Af, africa, Negro, blacks, black, af, frica, Blacks \\
\cmidrule(lr){2-4}
 & Latino & MLP.v$_{9242}^{32}$ & Spanish, Hispanic, Spanish, spanish, Mexican, Argentine, Santiago, Mexico, Puerto, Madrid, pesos, Chavez, Ecuador, Spain, Juan, Hispan, Colombian, Hispanics, Carlos, Chile \\
\cmidrule(lr){2-4}
 & \multirow{2}{*}{Native Am.} & MLP.v$_{6893}^{32}$ & colon, Colon, Colon, colon, colonial, Colonial, colonization, colonies, Colony, colony, Colonel, 殖, olon, icolon, Colin, OLON, kol, COL, -Col, colore \\
 \cmidrule{3-4}
 & & MLP.v$_{1186}^{31}$ & native, energy, Energy, native, energy, Native, Native, -native, Energy, natives, \_native, .native, \_energy, -energy, supply, 能源, culture, /native, nice, Higher \\
\bottomrule
\end{tabular}
}
\caption{Full top-20 tokens for race-associated neurons in Llama-3.1-8B (ToxiGen).
{\scriptsize Translations: 中国 (China), 中國 (China), 中国 (China), {\arabfont چین} (China), Çin (China), {\greekfont 중국} (China), 東 (East), 东 (East), Doğu (East), Đông (East), {\arabfont شرق} (East), vých (East), {\cyrfont вост} (East), 殖 (colonial/colonize), 能源 (energy).}}
\label{tab:appendix_toxigen_llama}
\end{table*}

\subsection{Probe Token Projections (C-REACT)}
\label{app:creact_tokens}
Table~\ref{tab:appendix_creact_direct} and Table~\ref{tab:appendix_creact_indirect} present the complete top-20 tokens projected by each race direction probe for direct and indirect mentions respectively.

\begin{table*}[ht]
\centering
\resizebox{\textwidth}{!}{%
\begin{tabular}{lll}
\toprule
\textbf{Model} & \textbf{Group} & \textbf{Top 20 Tokens} \\
\midrule
\multirow{3}{*}{Qwen2.5-7B}
& White & 白, generado, -Nazi, ksam, onn, pl, owl, ucas, *\&, '\#\{, .toByteArray, \_EXTENSIONS, avras, onFailure, 镪, ski, Nederland, 好运, Luft, Ski \\
\cmidrule(lr){2-3}
& Asian & Asian, Asian, Asians, Asia, 亚洲, Asia, asia, Asi, アジア, asiat, asian, 舢, Taiwan, Singapore, 舶, Singapore, Tai, Canton, Taiwanese, {\cyrfont ракти} \\
\cmidrule(lr){2-3}
& Black/AA & african, African, 非洲, -AA, Africans, Africa, ienda, Africa, {\arabfont السود}, 疟, mongo, /black, aina, AA, .BL, .Black, .Mongo, Afro, Nigerian, Black \\
\midrule
\multirow{3}{*}{Mistral-7B}
& White & ogle, ASC, cip, Kurt, heid, unächst, kle, þ, criptor, ór, NOP, och, zym, hid, eu, cow, cí, zens, vas, awa \\
\cmidrule(lr){2-3}
& Asian & Asian, Taiwan, Japanese, Hong, Malays, Japan, Singapore, Chinese, Malaysia, Pak, Indones, Korean, Tai, Asia, Philippines, Philipp, Tokyo, Tok, jap, Sri \\
\cmidrule(lr){2-3}
& Black/AA & African, Afr, Africa, blacks, Niger, Jama, Negro, Nigeria, Black, Af, ament, slavery, sist, black, external, external, Af, AA, lando, slave \\
\midrule
\multirow{3}{*}{Llama-3.1-8B}
& White & ithe, lan, Fres, ycz, hs, Wake, Bread, bread, itra, bairro, Fans, 1, Josh, wie, Jackets, ジオ, Marina, {\greekfont γι}, Josh, avan \\
\cmidrule(lr){2-3}
& Asian & Asian, Asian, Asians, asian, Indonesian, Asia, Taiwanese, Asia, asiat, Vietnamese, Japanese, Korean, Oriental, Malaysian, Buddhist, 亚洲, orean, Filipino, Chinese, Indones \\
\cmidrule(lr){2-3}
& Black/AA & black, African, Black, african, black, frican, Afro, /black, negro, Negro, 黑, blacks, BLACK, zwarte, 黑, Black, .Black, \_black, Africa, Africans \\
\bottomrule
\end{tabular}
}
\caption{Full top-20 probe token projections for C-REACT direct mentions (explicit race/ethnicity).
{\scriptsize Translations: 白 (white), generado (generated), 镪 (money), Nederland (Netherlands), 好运 (good luck), Luft (air), 亚洲 (Asia), {\greekfont アジア} (Asia), 舢 (Shantou), 舶 (ship), {\cyrfont ракти} (practice), 非洲 (Africa), {\arabfont السود} (black), 疟 (malaria), unächst (initially), þ (th), ór (or), cí (here), zens (citizens), bairro (neighborhood), ジオ (Geo), {\greekfont γι} (gi), 黑 (black), zwarte (black).}}
\label{tab:appendix_creact_direct}
\end{table*}

\begin{table*}[ht]
\centering
\resizebox{\textwidth}{!}{%
\begin{tabular}{lll}
\toprule
\textbf{Model} & \textbf{Group} & \textbf{Top 20 Tokens} \\
\midrule
\multirow{3}{*}{Qwen2.5-7B}
& White & RaisePropertyChanged, Russian, 俄罗斯, 橼, 圬, ocaly, Russia, \_backward, Kremlin, Russian, *)((, .defaultValue, RU, russian, egrator, ipsis, Rad, UCE, abis, ENCIL \\
\cmidrule(lr){2-3}
& Asian & .Dao, 华人, Chinese, 嵴, .insertBefore, Xia, chinese, ettel, Chinese, Tibetan, China, Wong, /apache, 华南, dao, dx, 갔, Jun, utenant, stylesheet \\
\cmidrule(lr){2-3}
& Black/AA & Hait, Haiti, Tropical, )\_\_, eneg, \%;", (\$., 埴, \%;">, 彗, tropical, 垓, |\`, .iterator, 热带, \%;">, ];//, loo, estring, ***** \\
\midrule
\multirow{3}{*}{Mistral-7B}
& White & Moscow, Russian, Ukrain, Ukraine, Russians, Polish, Ukr, Russia, vod, Soviet, Vlad, russ, Kaz, Bulgar, Lieutenant, Mik, icz, Roma, dou, Serge \\
\cmidrule(lr){2-3}
& Asian & Korea, Korean, Asian, trag, apis, Shan, Vietnam, gram, sg, apore, ga, Aires, Assembly, WD, lag, Nor, Viet, Schw, gan, cent \\
\cmidrule(lr){2-3}
& Black/AA & Caribbean, Jama, Braz, Af, Cuba, Niger, Nigeria, Cub, Bah, São, Core, Currency, Bras, island, Hy, hur, Curt, Af, Brazil, mont \\
\midrule
\multirow{3}{*}{Llama-3.1-8B}
& White & Russia, Kremlin, Russian, Russians, Putin, Moscow, Ukraine, Putin, Ukrainian, Russian, Russia, Ukrain, russian, Rus, Belarus, Rus, russe, russ, Kiev, Rusya \\
\cmidrule(lr){2-3}
& Asian & Cambodia, Asian, Chung, Cheng, Chinese, Camb, Kang, wang, Chinese, chinese, Buddhism, asian, Hong, Bangalore, Buddhist, Bang, Asian, Malaysia, Korean, {\cyrfont ерг} \\
\cmidrule(lr){2-3}
& Black/AA & Hait, Haiti, Caribbean, Maurit, Dominican, Bahamas, hait, Cre, Cre, Jama, ibbean, Trinidad, ingt, Jean, Cameroon, François, Maurice, Jean, {\cyrfont анка}, Santo \\
\bottomrule
\end{tabular}
}
\caption{Full top-20 probe token projections for C-REACT indirect mentions (language/country).
{\scriptsize Translations: 俄罗斯 (Russia), 橼 (yuan; Chinese monetary unit), 圬 (plaster), 华人 (ethnic Chinese), 嵴 (ridge), 华南 (South China), {\greekfont 갔} (went), 埴 (clay), 彗 (comet), 垓 (vast number), 热带 (tropical), São (Saint/São), Rusya (Russia), {\cyrfont ерг} (erg), {\cyrfont анка} (anka).}}
\label{tab:appendix_creact_indirect}
\end{table*}

\subsection{Race-Associated Neurons (C-REACT Direct)}
\label{app:creact_direct_neurons}
Table~\ref{tab:appendix_neurons_creact_direct} presents the complete list of race-associated neurons identified from C-REACT direct mentions (explicit race/ethnicity).

\begin{table*}[ht]
\centering
\resizebox{\textwidth}{!}{%
\begin{tabular}{llll}
\toprule
\textbf{Model} & \textbf{Group} & \textbf{Neuron} & \textbf{Top 20 Tokens} \\
\midrule
\multirow{19}{*}{Qwen2.5-7B}
& \multirow{8}{*}{White} 
& MLP.v$^{28}_{16880}$ & 英国, Dutch, French, 法国, 意大利, Italian, German, 荷兰, France, British, Belgian, European, Britain, Germany, 德国, Spanish, Italy, Spain, 欧洲, French \\
\cline{3-4}
& & MLP.v$^{27}_{17660}$ & German, 万欧元, ドイツ, 德国, German, Germany, Germans, EU, 荷兰, euro, €, 欧盟, Germany, Swiss, 欧元, Euro, Dutch, 瑞士, 欧, Switzerland \\
\cline{3-4}
& & MLP.v$^{25}_{4157}$ & UEFA, 地中海, Mediterranean, Cyprus, Pane, Roman, 欧盟, Euro, Roman, chalk, 希腊, Europe, UES, Rom, 罗马, EU, Kn, Europe, 橡, 騰 \\
\cline{3-4}
& & MLP.v$^{25}_{8669}$ & fod, ].', []>, sterdam, Ừ, Spain, Gaines, Spain, 比利时, ⊊, WRITE, 英国, europe, <!--[, ======, \%@", 示, 西班牙, \_IMPORTED, euros \\
\cmidrule(lr){2-4}
& \multirow{14}{*}{Asian}
& MLP.v$^{28}_{13406}$ & 在日本, Japanese, 日本, Japanese, Japan, Japan, 日本人, Tokyo, japan, 东京, 일본, 日军, japanese, japon, Nhật, Tok, Hiro, 日本の, Osaka, jap \\
\cline{3-4}
& & MLP.v$^{28}_{16570}$ & 习, 習, 惯, habit, 习惯, 的习惯, habitual, Habit, 習慣, habit, habits, accustomed, 习近平, 习惯了, 慣, 习近, 惯例, 习俗, 习近平总, 总书记 \\
\cline{3-4}
& & MLP.v$^{27}_{6943}$ & Asian, Asian, 亚洲, Asia, Chinese, Asia, Chinese, China, 印度, Indian, India, India, Asians, China, Indian, chinese, 中國, Oriental, Japanese, asian \\
\cline{3-4}
& & MLP.v$^{27}_{217}$ & Asian, 亚洲, Asia, Asian, Asians, Asia, asian, asia, アジア, 亚, 亞, Asi, asia, asiat, asi, 亚太, {\thaifont เอเช}, \_As, asi, AS \\
\cline{3-4}
& & MLP.v$^{26}_{5187}$ & Chinese, Italian, Chinese, Irish, Italian, Asian, Japanese, chinese, Mexican, Vietnamese, Korean, German, Greek, Asian, Portuguese, Latin, Indian, African, German, italian \\
\cline{3-4}
& & MLP.v$^{26}_{8828}$ & 亚洲, Asian, Asia, Asian, Latin, 欧, Asia, 亚太, 欧洲, 非洲, 东亚, Southeast, Europe, 拉丁, European, 东南亚, 美洲, アジア, Euras, Asians \\
\cline{3-4}
& & MLP.v$^{25}_{15029}$ & Chinese, China, China, Chinese, chinese, china, china, 中国, 中国的, -China, 중국, {\arabfont الصين}, 中國, Asian, 在中国, 是中国, 亚洲, 由中国, Asian, Asia \\
\cmidrule(lr){2-4}
& \multirow{16}{*}{Black/AA}
& MLP.v$^{28}_{11088}$ & racial, racially, racist, 种族, racism, racial, Harlem, segregated, segregation, Rac, interracial, Ethnic, Afro, hetto, ethnic, rac, Interracial \\
\cline{3-4}
& & MLP.v$^{28}_{10048}$ & µ, ctx, ctx, µ, (ctx, Ken, Ken, Ctx, Cape, \_ctx, .ctx, ctx, μ, \_CTX, 南非, context, CTX, Kenya, μ, context \\
\cline{3-4}
& & MLP.v$^{27}_{2240}$ & black, 黑, black, Black, Black, 黑色, -black, BLACK, BLACK, \_black, blacks, .black, /black, ブラック, .Black, ⿊, 黑白, \_BLACK, blacklist \\
\cline{3-4}
& & MLP.v$^{27}_{16596}$ & Ken, Ken, 肯, Nam, Liber, Bot, Tanz, Burk, Bur, Chad, ken, 赞, Nam, Kenya, Ug, Jordan, ken, Maur, Per, Jordan \\
\cline{3-4}
& & MLP.v$^{26}_{18261}$ & 黑, black, black, 黑色, Black, -black, Black, \_black, .black, dark, ⿊, BLACK, /black, 黑暗, blacks, .Black, BLACK, đen, darken \\
\cline{3-4}
& & MLP.v$^{26}_{1091}$ & 黑, black, 黑色, Black, Black, black, -black, BLACK, BLACK, /black, blacks, \_black, dark, Blacks, Dark, blacklist, 黑暗, ⿊, .Black \\
\cline{3-4}
& & MLP.v$^{25}_{10230}$ & Af, af, 非洲, Af, African, slave, AF, Africans, AF, african, slavery, 奴隶, 奴, slaves, Slave, slave, Afro, Slave, Africa, afr \\
\cline{3-4}
& & MLP.v$^{25}_{10739}$ & 非洲, African, Africa, Africa, Africans, african, Ghana, Nigerian, Nigeria, afr, Kenya, Nairobi, Niger, -Saharan, Nd, Tanzania, Uganda, africa, Nz, Lagos \\
\midrule
\multirow{13}{*}{Mistral-7B}
& \multirow{4}{*}{White}
& MLP.v$^{32}_{1606}$ & England, France, Europe, Switzerland, Britain, Holland, Spain, Australia, España, Austria, Germany, Denmark, Francia, Canada, Europa, America, Wales, Deutschland, Belgium, Sweden \\
\cline{3-4}
& & MLP.v$^{32}_{12760}$ & €, €, Finn, Mediterranean, Belgium, Italy, Finland, Belg, Luxem, Netherlands, Milan, Italian, Portugal, Denmark, Czech, Madrid, Norway, Sweden, Spain, Amsterdam \\
\cline{3-4}
& & MLP.v$^{32}_{9831}$ & European, Eu, Europe, Europe, EU, eu, Europ, Euro, europe, eu, Europa, europ, {\cyrfont Евро}, urope, UEFA, €, urop, Є, €, {\cyrfont вро} \\
\cline{3-4}
& & MLP.v$^{30}_{1521}$ & Pennsylvania, Michigan, Kansas, Philadelphia, Verm, ahl, Detroit, rus, {\cyrfont ники}, zel, Vic, Buc, ru, dispos, Hein, gotta, xa, una, anne, onic \\
\cmidrule(lr){2-4}
& \multirow{4}{*}{Asian}
& MLP.v$^{32}_{4453}$ & Japanese, Korean, Kol, Japan, Kor, Kaz, Sak, Taiwan, Korea, Nak, Kur, Kom, Bangl, Pakistan, Kab, Kon, Pak, Tibet, Ku, Asian \\
\cline{3-4}
& & MLP.v$^{31}_{2346}$ & Japanese, anime, Japan, Tokyo, jap, Jap, Korean, oji, apan, Tok, Asian, Taiwan, Israeli, Chinese, Korea, Indones, aku, Viet, Sak, oshi \\
\cline{3-4}
& & MLP.v$^{30}_{2137}$ & Wil, mal, Mal, Ker, Mal, mal, Malays, Malaysia, ker, <\textbackslash, \$\textbackslash, -\textbackslash, \$|\textbackslash, Carm, (\$\textbackslash, /\textbackslash, mals, -\textbackslash, +\textbackslash, >\textbackslash \\
\cline{3-4}
& & MLP.v$^{30}_{8986}$ & asis, asi, Ash, lic, ox, Asia, yk, ke, ash, ym, as, ass, Async, asia, ais, eper, ob, ash, ek \\
\cmidrule(lr){2-4}
& \multirow{5}{*}{Black/AA}
& MLP.v$^{32}_{5923}$ & Black, black, Black, black, blacks, 黑, 黒, Negro, BL, blk, African, BL, {\cyrfont чер}, Afr, Dark, {\cyrfont ч}, dark, {\cyrfont Чер}, {\cyrfont лек}, 블 \\
\cline{3-4}
& & MLP.v$^{31}_{8715}$ & African, Africa, Afr, Niger, Af, frica, Kenya, Af, Nigeria, fr, Negro, af, Afghan, blacks, FR, airo, Caribbean, Johannes, ₦, Jama \\
\cline{3-4}
& & MLP.v$^{30}_{3398}$ & African, Africa, Afr, Af, Af, Asian, fr, Niger, Aaron, Afghan, Kenya, frica, af, AF, Mexican, AF, European, Nigeria, af, Egyptian \\
\cline{3-4}
& & MLP.v$^{30}_{12572}$ & Black, black, Black, black, blacks, 黑, BL, 黒, BL, Negro, blk, {\cyrfont чер}, dark, African, Afr, 블, {\cyrfont лек}, dark, {\cyrfont Чер}, {\cyrfont ч} \\
\cline{3-4}
& & MLP.v$^{29}_{13186}$ & black, black, Black, blacks, Black, white, 黑, white, whites, African, 黒, Negro, 白, Afr, White, White, BL, {\cyrfont чер}, Af, brown \\
\midrule
\multirow{8}{*}{Llama-3.1-8B}
& \multirow{1}{*}{White}
& MLP.v$^{31}_{9094}$ & White, white, White, white, WHITE, -white, 白, WHITE, \_white, .White, whites, 白, .white, \_WHITE, Whites, beyaz, :white, trắng, {\arabfont سفید}, .WHITE \\
\cmidrule(lr){2-4}
& \multirow{3}{*}{Asian}
& MLP.v$^{32}_{5691}$ & Chinese, China, Chinese, China, chinese, china, -China, Beijing, 中国, Çin, 중국, Shanghai, 中國, china, {\arabfont چین}, 中国, Zhang, Jiang, Tencent, Guang \\
\cline{3-4}
& & MLP.v$^{31}_{14299}$ & Li, yuan, Dong, Huang, Liu, Wang, Chen, Yang, Tian, Zhou, Ding, Wu, dong, dong, Feng, wang, Zhang, Qin, Jiang, Guang \\
\cline{3-4}
& & MLP.v$^{29}_{5272}$ & Asia, Asia, Asian, Asian, asia, continent, Africa, continental, asian, Africa, asia, Asians, Continental, 亚洲, Asi, Europe, Continent, Latin, asi, Europe \\
\cmidrule(lr){2-4}
& \multirow{4}{*}{Black/AA}
& MLP.v$^{30}_{7195}$ & Mississippi, Jamaica, Jama, Caribbean, Louisiana, Trinidad, Haiti, LSU, Ghana, Hait, Baton, Harlem, Nigeria, negro, Negro, Jazz, Bahamas, Memphis, Nigerian, Zimbabwe \\
\cline{3-4}
& & MLP.v$^{30}_{3868}$ & black, Black, BLACK, Black, black, /black, BLACK, -black, 黑, blacks, \_black, .Black, blacklist, 黒, Blacks, .black, 黑, Blackburn, \_BLACK, blackout \\
\cline{3-4}
& & MLP.v$^{29}_{13826}$ & African, african, Africans, Africa, Africa, afr, Afro, Afr, frican, negro, Af, Afrika, Af, africa, Negro, blacks, black, af, frica, Blacks \\
\cline{3-4}
& & MLP.v$^{29}_{6824}$ & tropical, jungle, Tropical, Jungle, jung, ropical, trop, Fiji, Belize, mango, Congo, Jama, Caribbean, Hait, Honduras, Mango, BSON, Haiti, Safari, Jamaica \\
\bottomrule
\end{tabular}
}
\caption{Full list of race-associated neurons identified from C-REACT direct mentions.
{\scriptsize Translations: 英国 (UK), 法国 (France), 意大利 (Italy), 荷兰 (Netherlands), 德国 (Germany), 欧洲 (Europe), 万欧元 (ten-thousand euros), {\greekfont ドイツ} (Germany), 欧盟 (EU), 欧元 (euro), 瑞士 (Switzerland), 欧 (Europe), 地中海 (Mediterranean), 希腊 (Greece), 罗马 (Rome), 橡 (oak), 騰 (soar), Ừ (yes), 比利时 (Belgium), ⊊ (subsetneq), 示 (show), 西班牙 (Spain), 在日本 (in Japan), 日本 (Japan), 日本人 (Japanese), 东京 (Tokyo), {\greekfont 일본} (Japan), 日军 (Japanese army), Nhật (Japan), 日本の (Japan), 习 (Xi), 習 (Xi), 惯 (habit), 习惯 (habit), 的习惯 (habit), 習慣 (habit), 习近平 (Xi Jinping), 习惯了 (used to), 慣 (habit), 习近 (Xi-), 惯例 (convention), 习俗 (custom), 习近平总 (Xi Jinping), 总书记 (general secretary), 亚洲 (Asia), 印度 (India), 中國 (China), 亚 (Asia), 亞 (Asia), 亚太 (Asia-Pacific), {\thaifont เอเช} (Asia), 东亚 (East Asia), 欧洲 (Europe), 非洲 (Africa), 拉丁 (Latin), 东南亚 (Southeast Asia), 美洲 (Americas), {\greekfont アジア} (Asia), 中国 (China), 中国的 (China), {\greekfont 중국} (China), {\arabfont الصين} (China), 在中国 (China), 是中国 (China), 由中国 (China), 种族 (race), 南非 (South Africa), 黑 (black), 黑色 (black), {\greekfont ブラック} (black), ⿊ (black), 黑白 (black-and-white), 肯 (Ken), 赞 (praise), 黑暗 (dark), đen (black), 非洲 (Africa), 奴隶 (slave), 奴 (slave), {\cyrfont Евро} (Euro), Є (Euro), {\cyrfont вро} (Euro), {\cyrfont ники} (niki), 黒 (black), {\cyrfont чер} (black), {\cyrfont ч} (ch), {\cyrfont Чер} (black), {\cyrfont лек} (lek), 블 (black), 白 (white), beyaz (white), trắng (white), {\arabfont سفید} (white), 中国 (China), 中國 (China), {\arabfont چین} (China), Çin (China), {\greekfont 중국} (China), 亚洲 (Asia).}
}
\label{tab:appendix_neurons_creact_direct}
\end{table*}

\subsection{Race-Associated Neurons (C-REACT Indirect)}
\label{app:creact_indirect_neurons}
Table~\ref{tab:appendix_neurons_creact_indirect} presents the complete list of race-associated neurons identified from C-REACT indirect mentions (language/country).

\begin{table*}[ht]
\centering
\resizebox{\textwidth}{!}{%
\begin{tabular}{llll}
\toprule
\textbf{Model} & \textbf{Group} & \textbf{Neuron} & \textbf{Top 20 Tokens} \\
\midrule
\multirow{17}{*}{Qwen2.5-7B}
& \multirow{8}{*}{White} 
& MLP.v$^{28}_{8780}$ & Maryland, Pennsylvania, Baltimore, Ohio, Wisconsin, Maine, Minnesota, Illinois, Pittsburgh, Iowa, Ohio, Connecticut, Philadelphia, Chicago, Michigan, Milwaukee, Nebraska, Detroit, Seattle, 江苏省 \\
\cline{3-4}
& & MLP.v$^{28}_{9988}$ & 以色列, Israel, Jerusalem, Israel, Israeli, Noah, Hebrew, Zion, Moses, -Israel, Palestine, Israeli, Israelis, Biblical, biblical, Zionist, Luke, Palestinian, Jer, Nathan \\
\cline{3-4}
& & MLP.v$^{28}_{4318}$ & Ohio, Columbus, sz, Cleveland, Ohio, (sz, sz, 浙江, 杭州, Sz, Sz, 浙, 浙江省, 杭州市, SZ, Cincinnati, 杭, \_sz, Akron, Croatian \\
\cline{3-4}
& & MLP.v$^{27}_{17660}$ & German, 万欧元, ドイツ, 德国, German, Germany, Germans, EU, 荷兰, euro, €, 欧盟, Germany, Swiss, 欧元, Euro, Dutch, 瑞士, 欧, Switzerland \\
\cline{3-4}
& & MLP.v$^{26}_{3012}$ & 犹, Jew, Jewish, 猶, Jews, Judaism, {\cyrfont ев}, jewish, Juda, Hebrew, {\cyrfont евр}, 以色列, Israel, kosher, synagogue, -J, ewish, jew, Rabbi, Israeli \\
\cline{3-4}
& & MLP.v$^{26}_{3382}$ & 俄, Rus, Russ, RU, rus, Russians, Russ, Russia, Russian, russe, 俄罗斯, russ, Russo, 俄国, 莫斯科, Mos, Russia, Moscow, -Russian, RU \\
\cline{3-4}
& & MLP.v$^{25}_{4157}$ & UEFA, 地中海, Mediterranean, Cyprus, Pane, Roman, 欧盟, Euro, Roman, chalk, 希腊, Europe, UES, Rom, 罗马, EU, Kn, Europe, 橡, 騰 \\
\cline{3-4}
& & MLP.v$^{25}_{5123}$ & Ukraine, 乌, 乌克兰, Russia, Ukrainian, Russian, 俄罗斯, Ukrain, 烏, 乌克, Russia, war, Russian, -Russian, 冲突, russian, Russo, Conflict, conflict, 地 \\
\cmidrule(lr){2-4}
& \multirow{7}{*}{Asian}
& MLP.v$^{28}_{13406}$ & 在日本, Japanese, 日本, Japanese, Japan, Japan, 日本人, Tokyo, japan, 东京, 일본, 日军, japanese, japon, Nhật, Tok, Hiro, 日本の, Osaka, jap \\
\cline{3-4}
& & MLP.v$^{27}_{6943}$ & Asian, Asian, 亚洲, Asia, Chinese, Asia, Chinese, China, 印度, Indian, India, India, Asians, China, Indian, chinese, 中國, Oriental, Japanese, asian \\
\cline{3-4}
& & MLP.v$^{27}_{229}$ & Mã, Opcode, (equalTo, Singapore, Singapore, ').", wię, (defvar, ,tp, {\cyrfont Ла}, {\cyrfont Ново}, yne, (ls, Tah, 海南, établissement, airobi, Î, 槟 \\
\cline{3-4}
& & MLP.v$^{26}_{9908}$ & {\arabfont الثال}, {\arabfont الإثن}, Malays, Indones, protester, HttpMethod, Zambia, {\arabfont الجها}, Rodrig, Taiwanese, 辇, Jama, 加工厂, {\arabfont الاستث}, Karnataka, UserDao, Rousse, Böl, african \\
\cline{3-4}
& & MLP.v$^{26}_{13889}$ & chop, Conf, Mand, Canton, 功夫, mand, Dragon, Chop, Dragon, Conf, mand, dragon, Cant, dragon, Fu, Portsmouth, Bruce, Cath, 洗衣, Fuk \\
\cline{3-4}
& & MLP.v$^{25}_{2001}$ & 越, Vietnam, 越南, Viet, Nguyen, Viet, Vietnamese, .vn, Ho, Ph, Vu, Tran, viet, 就越, Ph, Dien, 越大, 越好, anh, gia \\
\cline{3-4}
& & MLP.v$^{25}_{15029}$ & Chinese, China, China, Chinese, chinese, china, china, 中国, 中国的, -China, 중국, {\arabfont الصين}, 中國, Asian, 在中国, 是中国, 亚洲, 由中国, Asian, Asia \\
\cmidrule(lr){2-4}
& \multirow{2}{*}{Black/AA}
& MLP.v$^{25}_{10230}$ & Af, af, 非洲, Af, African, slave, AF, Africans, AF, african, slavery, 奴隶, 奴, slaves, Slave, slave, Afro, Slave, Africa, afr \\
\cline{3-4}
& & MLP.v$^{25}_{10739}$ & 非洲, African, Africa, Africa, Africans, african, Ghana, Nigerian, Nigeria, afr, Kenya, Nairobi, Niger, -Saharan, Nd, Tanzania, Uganda, africa, Nz, Lagos \\
\midrule
\multirow{5}{*}{Mistral-7B}
& \multirow{4}{*}{White}
& MLP.v$^{32}_{2399}$ & Russian, Vlad, Moscow, Soviet, Russia, Ukrain, Russians, Ukraine, Czech, Ukr, Bulgar, Slov, vod, Cz, Aleks, Serge, sov, russ, Polish, Stalin \\
\cline{3-4}
& & MLP.v$^{31}_{7356}$ & Russell, Rus, Russ, Russian, rus, Russia, russ, Russians, Soviet, Moscow, {\cyrfont рус}, rust, rus, rust, {\cyrfont ру}, sov, Rug, Rud, ruin, Ruth \\
\cline{3-4}
& & MLP.v$^{30}_{4487}$ & Russell, Russ, russ, Russian, Rus, Russia, Russians, rus, rus, Moscow, {\cyrfont рус}, {\cyrfont Росси}, {\cyrfont России}, uss, Soviet, ussia, ussian, Vlad, USS, sov \\
\cline{3-4}
& & MLP.v$^{29}_{260}$ & Italian, Ital, Italy, ital, italien, Italia, italiano, ital, Giovanni, Francesco, Carlo, Gian, Sic, Gi, IT, Milan, Serie, acci, Giov, IT \\
\cmidrule(lr){2-4}
& \multirow{1}{*}{Black/AA}
& MLP.v$^{31}_{8715}$ & African, Africa, Afr, Niger, Af, frica, Kenya, Af, Nigeria, fr, Negro, af, Afghan, blacks, FR, airo, Caribbean, Johannes, ₦, Jama \\
\midrule
\multirow{10}{*}{Llama-3.1-8B}
& \multirow{4}{*}{White}
& MLP.v$^{32}_{10606}$ & Russian, Russians, Russia, Russian, Moscow, Soviet, Russia, -Russian, Putin, russ, russian, Kremlin, 俄, russe, Russ, USSR, Russell, .ru, Vladimir, Sergei \\
\cline{3-4}
& & MLP.v$^{32}_{10409}$ & Baltic, Sloven, Celtic, Gron, Flem, Slovenia, Austria, Sch, Luxembourg, Monaco, Gros, Croatia, Naples, Blond, Croatian, Baron, Austrian, Mont, Trent, Malta \\
\cline{3-4}
& & MLP.v$^{31}_{12584}$ & Russell, Russ, Russ, Rus, RUS, Rus, russ, rus, Russo, Russia, Russian, Rusya, 俄, Russians, {\cyrfont Рус}, -Russian, rus, Russia, russe, Russian \\
\cline{3-4}
& & MLP.v$^{29}_{4193}$ & Czech, Hungarian, Boh, Slovak, Budapest, Hungary, Hung, Slovakia, Prague, Poland, polish, Polish, Sloven, Croatian, Krak, lovak, Brno, Slovenia, Romania, Croatia \\
\cmidrule(lr){2-4}
& \multirow{3}{*}{Asian}
& MLP.v$^{32}_{5691}$ & Chinese, China, Chinese, China, chinese, china, -China, Beijing, 中国, Çin, 중국, Shanghai, 中國, china, {\arabfont چین}, 中国, Zhang, Jiang, Tencent, Guang \\
\cline{3-4}
& & MLP.v$^{32}_{6950}$ & Bhar, Bh, Sinh, Nagar, Shah, Muk, Hindi, Allah, Tamil, Gujar, Bollywood, Kh, Dh, Gujarat, Punjab, Kh, Mumbai, Uttar, Maharashtra, Jain \\
\cline{3-4}
& & MLP.v$^{29}_{10616}$ & Indian, Indian, Indians, India, indian, India, india, Ấn, Bollywood, {\arabfont الهند}, Indi, Indianapolis, Mumbai, ₹, Hindu, Bombay, Hindi, Hindus, 印, Delhi \\
\cmidrule(lr){2-4}
& \multirow{3}{*}{Black/AA}
& MLP.v$^{30}_{7195}$ & Mississippi, Jamaica, Jama, Caribbean, Louisiana, Trinidad, Haiti, LSU, Ghana, Hait, Baton, Harlem, Nigeria, negro, Negro, Jazz, Bahamas, Memphis, Nigerian, Zimbabwe \\
\cline{3-4}
& & MLP.v$^{29}_{6824}$ & tropical, jungle, Tropical, Jungle, jung, ropical, trop, Fiji, Belize, mango, Congo, Jama, Caribbean, Hait, Honduras, Mango, BSON, Haiti, Safari, Jamaica \\
\cline{3-4}
& & MLP.v$^{29}_{13826}$ & African, african, Africans, Africa, Africa, afr, Afro, Afr, frican, negro, Af, Afrika, Af, africa, Negro, blacks, black, af, frica, Blacks \\
\bottomrule
\end{tabular}
}
\caption{Full list of race-associated neurons identified from C-REACT indirect mentions (language/country).
{\scriptsize Translations: 江苏省 (Jiangsu), 以色列 (Israel), 万欧元 (ten-thousand euros), {\greekfont ドイツ} (Germany), 德国 (Germany), 荷兰 (Netherlands), 欧盟 (EU), 欧元 (euro), 瑞士 (Switzerland), 欧 (Europe), 犹 (Jew), 猶 (Jew), {\cyrfont ев} (Jew), {\cyrfont евр} (Jew), 俄 (Russia), 俄罗斯 (Russia), 俄国 (Russia), 莫斯科 (Moscow), 地中海 (Mediterranean), 希腊 (Greece), 罗马 (Rome), 橡 (oak), 騰 (soar), 乌 (Ukraine), 乌克兰 (Ukraine), 烏 (Ukraine), 乌克 (Ukraine), 冲突 (conflict), 地 (land), 在日本 (in Japan), 日本 (Japan), 日本人 (Japanese), 东京 (Tokyo), {\greekfont 일본} (Japan), 日军 (Japanese army), Nhật (Japan), 日本の (Japan), 亚洲 (Asia), 印度 (India), 中國 (China), Mã (code), wię (bond), {\cyrfont Ла} (La), {\cyrfont Ново} (Novo), 海南 (Hainan), 槟 (Penang), Î (I), {\arabfont الثال} (Tue), {\arabfont الإثن} (Mon), {\arabfont الجها} (jihad), {\arabfont الاستث} (exception), 辇 (carriage), 加工厂 (factory), Böl (region), 功夫 (kung fu), 洗衣 (laundry), 越 (Viet), 越南 (Vietnam), 就越 (then more), 越大 (bigger), 越好 (better), 中国 (China), 中国的 (China), {\greekfont 중국} (China), {\arabfont الصين} (China), 在中国 (China), 是中国 (China), 由中国 (China), 非洲 (Africa), 奴隶 (slave), 奴 (slave), {\cyrfont рус} (Russian), {\cyrfont ру} (ru), {\cyrfont Росси} (Russia), {\cyrfont России} (Russia), italiano (Italian), italien (Italian), Italia (Italy), Ấn (India), {\arabfont الهند} (India), 印 (India).}
}
\label{tab:appendix_neurons_creact_indirect}
\end{table*}

\subsection{Example Model Prediction Outputs}
\label{sec:example_outputs}

Figure~\ref{fig:example_outputs} presents example model outputs after neuron intervention. Correct predictions result in valid racial category labels, while failures to provide a valid label are classified as Unknown.

\begin{figure}[ht]
\centering
\resizebox{\columnwidth}{!}{%
\begin{tabular}{p{2.5cm}p{9cm}}
\toprule
\textbf{Category} & \textbf{Model Output} \\
\midrule
\multicolumn{2}{l}{\textit{Correct predictions (White)}} \\
\midrule
Example 1 & [White] You are an AI assistant. Provide \\
Example 2 & [White] You are an AI assistant. User \\
\midrule
\multicolumn{2}{l}{\textit{Unknown outputs (invalid or malformed)}} \\
\midrule
Example 1 & [Russian] The race or ethnicity that best fits \\
Example 2 & [Russian] Based on the information provided in the \\
Example 3 & [Yellow] [X] [Black or African \\
Example 4 & [Yellow] The provided options do not include " \\
\bottomrule
\end{tabular}
}
\caption{Example outputs after Direct neuron intervention. Correct predictions produce valid category labels (top), while unstable interventions at high amplification factors produce invalid outputs classified as Unknown (bottom).}
\label{fig:example_outputs}
\end{figure}

\subsection{Implementation Details and Package Settings}

We implement model loading, inference, and activation extraction using PyTorch and Hugging Face Transformers. Unless otherwise specified, we use the default inference settings provided by the corresponding model checkpoints. Linear probes are trained on hidden states extracted from the final layer residual stream, averaged across token positions. For neuron identification, we compute cosine similarity between MLP value vectors and the learned probe directions, and inspect the top projected tokens using the model unembedding matrix. For intervention experiments, we use PyTorch forward hooks on the MLP down projection and apply amplification factors of 5, 10, and 20, as described in Section~6.3. Code and configuration files for reproducing the main analyses are available at \url{https://github.com/LARK-NLP-Lab/LLM-Bias-Interpretability}.

\end{document}